\documentclass{article} %
\usepackage[final]{colm2026_conference}  %

\usepackage[T1]{fontenc}    %
\usepackage{algorithm,algorithmicx}
\usepackage[noend]{algpseudocode}
\usepackage{amsmath,amssymb,amsfonts,amsthm,mathtools}
\usepackage{bm,bbm}
\usepackage{duckuments}
\usepackage{enumitem}
\usepackage{graphicx}
\usepackage{float}
\usepackage[nice]{nicefrac}
\usepackage[multiple]{footmisc}
\usepackage{caption,subcaption}
\usepackage{cprotect}  %
\usepackage{textcomp}
\usepackage{stfloats}
\usepackage{hyperref}
\usepackage[capitalize,noabbrev]{cleveref}
\usepackage{listings}
\usepackage{lipsum}
\usepackage{longtable,tabularx,booktabs,wrapfig}
\usepackage{makecell}
\usepackage{multirow}
\usepackage{siunitx}
\usepackage{url}
\usepackage[dvipsnames]{xcolor}
\usepackage{xspace}
\usepackage[utf8]{inputenc}
\usepackage{pgfplots}
\pgfplotsset{compat=1.18}
\usepackage{tcolorbox,tabularray}
\tcbuselibrary{breakable}
\usepackage{fancyvrb}
\usepackage{placeins}
\usepackage{colortbl}
\usepackage{microtype}

\usepackage{lineno}

\definecolor{darkblue}{rgb}{0, 0, 0.5}
\definecolor{cadmiumgreen}{rgb}{0.0, 0.42, 0.24}
\definecolor{cornellred}{rgb}{0.7, 0.11, 0.11}
\hypersetup{
  urlcolor=MidnightBlue,
  linkcolor=cornellred,
  citecolor=MidnightBlue,
  colorlinks=true,
}

\newcommand{\mname}{VPS}

\title{Correct Answers from Sound Reasoning: \\ Verifiable Process Supervision for Language Models}

\author{\textbf{Kyuyoung Kim\textsuperscript{1}}\; 
    \textbf{Kevin Wang\textsuperscript{2}}\;
    \textbf{Yunfei Xie\textsuperscript{3}}\;
    \textbf{Peiyang Xu\textsuperscript{4}}\;
    \textbf{Peiyao Sheng\textsuperscript{6}} \\
    \textbf{Chen Wei\textsuperscript{3}}\;
    \textbf{Zhangyang Wang\textsuperscript{2}}\;
    \textbf{Jinwoo Shin\textsuperscript{1}}\;
    \textbf{Pramod Viswanath\textsuperscript{4,6}}\;
    \textbf{Sewoong Oh\textsuperscript{5,6}} \\
    \textsuperscript{1}KAIST AI\; \textsuperscript{2}University of Texas, Austin\;
    \textsuperscript{3}Rice University\; \\ \textsuperscript{4}Princeton University\;
    \textsuperscript{5}University of Washington\; \textsuperscript{6}Sentient Labs
}

\begin{document}

\ifcolmsubmission
\linenumbers
\fi

\maketitle

\begin{abstract}

Training language models to produce both correct answers and sound reasoning remains an open challenge.
Reinforcement learning with verifiable rewards typically optimizes only final outcomes, which can improve task accuracy at the expense of reasoning quality, producing inaccurate, incomplete, or inconsistent traces.
We propose verifiable process supervision ({\mname}), a post-training framework that jointly optimizes prediction accuracy and reasoning quality by supervising structured intermediate claims.
We first apply supervised fine-tuning to induce a structured reasoning format, enabling deterministic extraction and verification of intermediate claims for process-level rewards.
To address the heterogeneous difficulty of reasoning subtasks, we introduce adaptive weighting that prioritizes components with the largest remaining errors, creating an implicit curriculum.
We evaluate {\mname} on chess as a controlled testbed where reasoning steps can be deterministically verified against engine signals.
While outcome-only RL improves move accuracy, it sharply degrades reasoning quality, increasing win-rate error by up to 112\% and reducing internal consistency by up to 69\%.
In contrast, {\mname} preserves accuracy while significantly improving reasoning quality, reducing win-rate error by up to 30\% and restoring consistency to near saturation.
A reasoning-space analysis further shows that, without a structured prior, outcome-only RL converges to budget-dependent shortcuts rather than sound multi-step reasoning.
Beyond chess, we observe the same phenomenon on math reasoning, where outcome-only RL improves accuracy while degrading step-level arithmetic and consistency, whereas {\mname} maintains both.
These results show that {\mname} enables language models to reason both accurately and reliably in verifiable domains.

\end{abstract}

\section{Introduction}
\label{s:intro}

Language models have recently shown strong performance on tasks that require multi-step reasoning, including mathematical problem solving, code generation, and game playing~\citep{shao2024deepseekmath,ma2025mixing}.
Much of this progress has been driven by reasoning-oriented post-training methods such as reinforcement learning with verifiable rewards (RLVR; \citealt{guo2025deepseek}) and reasoning distillation~\citep{cetin2025reinforcement,guha2025openthoughts}.
In these approaches, models are typically trained to generate explicit reasoning traces alongside their final predictions, both to improve performance and to expose a channel through which their decisions can be inspected.

A central difficulty, however, is that high task accuracy does not guarantee high-quality reasoning.
In RLVR, optimization is typically driven by the correctness of the final answer, leaving intermediate reasoning unconstrained by the optimization objective.
As a result, models can produce better predictions while their exposed reasoning becomes incomplete, inaccurate, or internally inconsistent. 
Similar discrepancies between task performance and reasoning quality have been observed in related settings~\citep{shao2025deepseekmath}.
This failure mode is particularly concerning when the reasoning trace itself is an object of interest, such as for debugging, auditing, or downstream reuse.
One response is to directly supervise reasoning using external evaluators such as LLM judges~\citep{hu2026rewarding}, but these approaches are computationally expensive and can introduce noise into the training signal.

We present a simple alternative: \textbf{verifiable process supervision ({\mname})}, a post-training framework for verifiable domains that rewards both final outcomes and intermediate reasoning claims that can be deterministically verified against ground truth.
Specifically, we first apply supervised fine-tuning (SFT) on synthetically constructed traces to induce a structured reasoning format.
This structure enables reliable extraction of intermediate claims using simple syntactic rules rather than semantic parsing.
We then evaluate these claims against domain constraints and ground-truth signals to compute process-level rewards, which are aggregated into a reasoning reward and combined with the standard prediction accuracy reward during RL training.
To address the heterogeneous difficulty of reasoning subtasks, we further introduce adaptive weighting that allocates greater weight to components with the largest remaining errors, effectively inducing a curriculum over reasoning subskills. 
This combined objective encourages the model to both produce correct outputs and generate reasoning traces consistent with the underlying state.

\begin{figure}[t]
    \small
    \centering
    \includegraphics[width=0.95\linewidth]{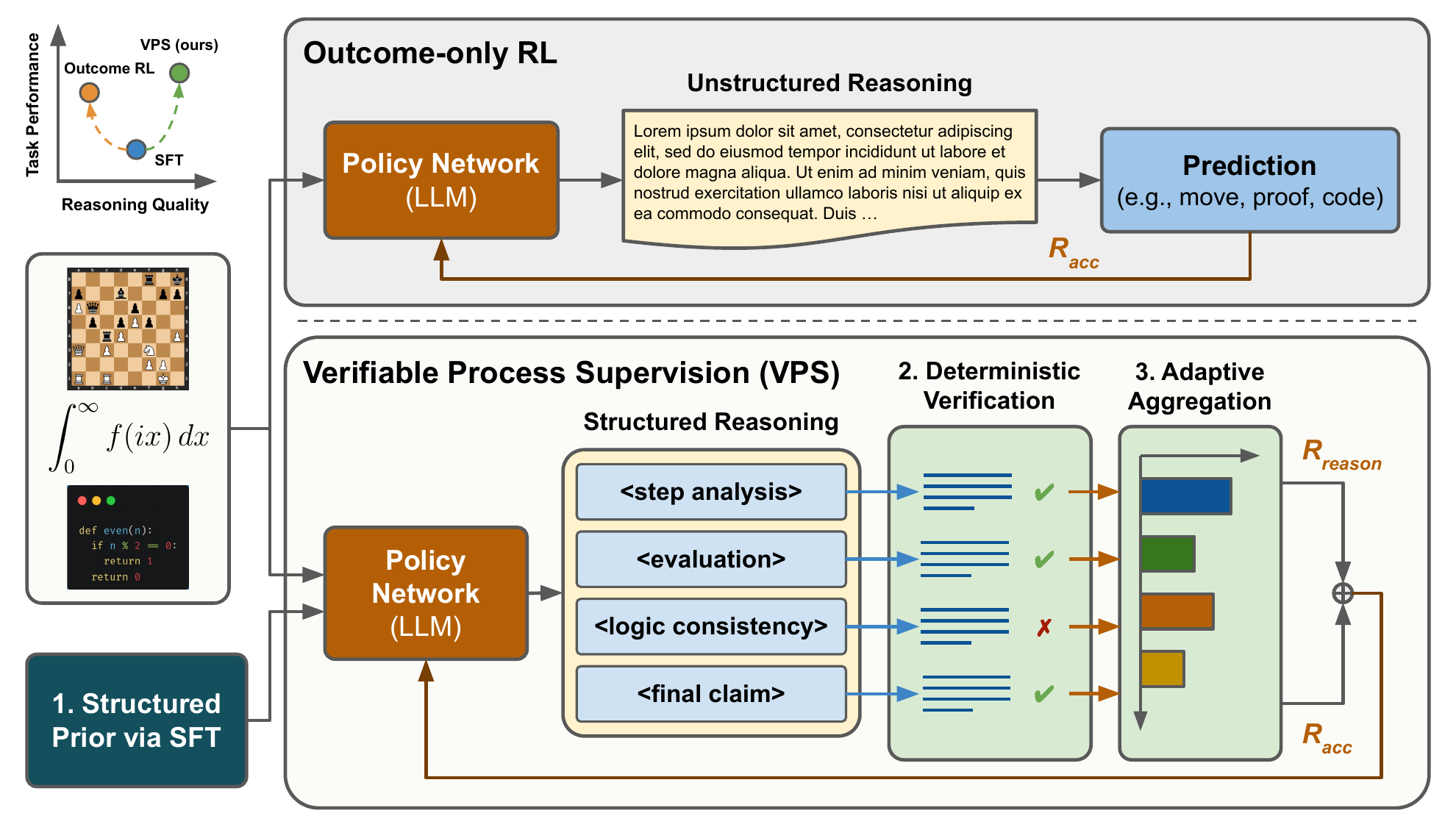}
    \vspace{-5pt}
    \caption{\textbf{Overview of {\mname}.} (1) A structured reasoning prior is induced via supervised fine-tuning, enabling syntactic extraction of intermediate claims. (2) During RL, these claims are verified against ground-truth signals to produce process-level rewards. (3) The rewards are adaptively weighted based on subtask performance, focusing learning on components with the largest remaining errors and inducing a curriculum over reasoning subskills.}
    \label{fig:concept}
    \vspace{-18pt}
\end{figure}

To assess whether {\mname} improves reasoning quality without sacrificing task performance, we evaluate it in chess as a controlled testbed for verifiable reasoning.
Chess offers exact rules, strong engine-based reference signals, and interpretable components such as candidate selection and position evaluation, enabling fine-grained verification of multi-step reasoning.
Across four open reasoning models, outcome-only GRPO improves prediction accuracy but substantially degrades reasoning quality, increasing win-rate error and reducing internal consistency.
In contrast, {\mname} preserves these accuracy gains while markedly improving reasoning quality, reducing win-rate error and bringing consistency close to saturation (Table~\ref{table:main-table}). 
Adaptive weighting further improves upon fixed aggregation, particularly on subtasks that are inherently difficult for language models.
Further analysis shows that, without a structured prior, outcome-only RL converges to budget-dependent shortcut strategies rather than principled multi-step reasoning.
Beyond this controlled setting, we observe the same phenomenon on GSM8K math reasoning: outcome-only RL improves answer accuracy while degrading intermediate arithmetic correctness and consistency, whereas {\mname} preserves accuracy and substantially improves reasoning quality.
Overall, these results suggest that deterministic verification of intermediate reasoning, combined with an adaptive curriculum over subskills, is an effective approach for training models that achieve strong task performance with accurate and consistent reasoning.

In summary, our main contributions are as follows:
\begin{itemize}[leftmargin=7mm]
    \item \textbf{Empirical characterization of outcome-only RL.} Through extensive experiments, we show that outcome-only RL improves task performance while degrading reasoning quality, revealing a systematic decoupling between the two.
    \item \textbf{Verifiable reasoning supervision.} We introduce a post-training framework that enables scalable process supervision through deterministic verification of structured intermediate claims, jointly optimizing prediction accuracy and reasoning quality.
    \item \textbf{Evaluation across verifiable reasoning domains.} Using chess as a fine-grained testbed and GSM8K as a second verifiable domain, we show that {\mname} preserves accuracy while substantially improving reasoning quality across model families and tasks.
\end{itemize}

\section{Related work}
\label{s:related}

\paragraph{Process supervision for reasoning models.}

In RL post-training of reasoning models, rewards are often assigned only based on final-answer correctness (e.g., exact match in math)~\citep{guo2025deepseek,olmo2025olmo}.
A key limitation of this outcome-only supervision is that it can assign high rewards even when generated reasoning traces are flawed~\citep{shao2025deepseekmath}.
While process supervision addresses this issue by incorporating feedback on intermediate reasoning~\citep{uesato2022solving,lightman2023lets,wang2024math}, existing approaches often rely on learned process reward models~\citep{luo2024improve} or LLM-based judges~\citep{zhang2025lessons}, introducing additional cost and potential noise.
When reasoning decomposes into multiple subtasks of varying difficulty, balancing their contributions during training becomes an additional challenge.

In this work, we propose \emph{verifiable process supervision}, which leverages structured reasoning priors and deterministic verification of intermediate claims to provide scalable process supervision without learned reward models or LLM judges.
We further introduce adaptive subtask weighting that prioritizes components with the greatest remaining headroom, inducing an implicit curriculum~\citep{bengio2009curriculum} that improves training efficiency.
Concurrently, \cite{pronesti2026beyond} propose verifiable process reward models based on step-level verification in structured tasks; in contrast, we study verifiable domains where intermediate reasoning structure is not given a priori, using SFT-induced structure and adaptive weighting to enable scalable verification.
A related line verifies reasoning through formal intermediates checked by a theorem prover~\citep{chen2026learning}; in contrast, {\mname} operates on multi-component reasoning where intermediate reasoning is not naturally expressed in a formally verifiable language, instead inducing structured reasoning through SFT so that intermediate claims can be extracted and verified against domain signals.

\paragraph{Language models and games.}

Language models have recently been explored as agents in a variety of game environments~\citep{yao2025spin,guertler2025textarena,xie2026memo}.
With appropriate prompting and agent designs, LLMs have demonstrated meaningful capabilities in interactive games~\citep{hu2024pokellmon,ma2024large,madge2024large}.
Within this broader landscape, chess provides a particularly suitable setting for studying verifiable reasoning.
Prior work applies LLMs to chess tasks including move prediction~\citep{noever2020chess}, state tracking~\citep{toshniwal2022chess}, and policy learning~\citep{feng2024chessgpt,zhang2025complete,hwang2025can}.
However, these approaches largely treat chess as a prediction problem and do not explicitly supervise intermediate reasoning.
In this work, we use chess as a testbed for verifiable reasoning, introducing a framework that transforms free-form reasoning into verifiable intermediate claims and adaptively supervises reasoning subtasks to jointly optimize task performance and reasoning quality.

\section{Adaptive verifiable process supervision}
\label{s:method}

We introduce a framework that jointly optimizes prediction accuracy and reasoning quality in verifiable domains.
First, we apply SFT on synthetically constructed traces to instill a \textit{structured reasoning prior} in the model, encouraging it to generate outputs in a predictable format amenable to automated verification.
Second, we use a reasoning verifier that extracts individual claims from the structured trace and evaluates them against ground-truth signals, producing a set of subtask rewards.
Finally, we aggregate these rewards using adaptive weighting, which dynamically adjusts aggregation weights based on the model's current performance on each subtask.
The resulting reasoning reward is combined with the standard prediction accuracy reward during RL training.
In summary, the three components address complementary challenges: SFT provides a verifiable reasoning interface, the verifier provides reliable process-level signals, and adaptive weighting allocates rewards according to the model's evolving capabilities.

Formally, let $x$ denote an input, $y^{*}$ the ground truth, and $\tau$ the model's reasoning trace generated prior to the final prediction $\hat{y}$.
The total reward is
\begin{equation}
    R(x, \tau, \hat{y}) \; = \; R_{\text{form}}(\hat{y}) + R_{\text{acc}}(\hat{y}, y^{*}) + \lambda \cdot R_{\text{reason}}(x, \tau),
\end{equation}
where $R_{\text{form}}$ and $R_{\text{acc}}$ are the standard rewards for format adherence and prediction accuracy. 
$R_{\text{reason}}$ is the reasoning reward that evaluates the correctness of reasoning claims extracted from $\tau$,
where $\lambda$ balances prediction accuracy and reasoning quality.

\begin{algorithm}[t]
\caption{RL Training with Verifiable Process Supervision}
\label{alg:rl_reasoning_training}

\begin{algorithmic}[1]

\Require Dataset $\mathcal{D}$, verifier $\mathcal{V}$, reasoning components $\{1,\dots,K\}$,
EMA decay $\alpha$, temperature $T$, min weight $w_{\min}$, reward balance $\lambda>0$, training steps $S$

\State Initialize policy $\pi$ via SFT on structured reasoning traces
\State Initialize EMA performance estimates, e.g., $\mu_k \leftarrow 0.5$ for all $k = 1,\dots,K$

\For{$t=1,\dots,S$}

    \State Sample batch of inputs $\{x_i\}_{i=1}^B \sim \mathcal{D}$
    \State Generate traces and predictions $\{(\tau_i,\hat{y}_i) \sim \pi(\cdot \mid x_i)\}_{i=1}^B$

    \For{$i=1,\dots,B$}
        \State Extract claims $\{c_{i,k}\}_{k=1}^K$ from $\tau_i$ and compute subtask rewards $\{r_{i,k} \leftarrow \mathcal{V}(c_{i,k})\}_{k=1}^K$
        \State Compute reasoning reward $R_{\text{reason},i} \leftarrow \sum_k w_k r_{i,k}$
        \State Compute total reward $R_i \leftarrow R_{\text{form},i} + R_{\text{acc},i} + \lambda R_{\text{reason},i}$
    \EndFor

    \State Update $\pi$ using an RL algorithm of choice with rewards $\{R_i\}_{i=1}^B$

    \State Compute batch mean $\bar{r}_k \leftarrow \frac{1}{B} \sum_{i=1}^B r_{i,k}$
    \State Update EMA $\mu_k \leftarrow (1-\alpha)\mu_k + \alpha \bar{r}_k$
    \State Update weights $w_k \leftarrow \text{normalize}\left(\max\left(\text{softmax}((1-\mu_k) / T)\right), w_{\min}\right)$

\EndFor

\end{algorithmic}
\end{algorithm}

\subsection{Structured reasoning prior via SFT}
\label{ss:sft-reason-prior}

A central challenge in assigning rewards to reasoning traces is that unstructured reasoning is difficult to parse and verify automatically, often requiring semantic evaluation from an LLM judge that is expensive and noisy.
We address this by first fine-tuning the model on synthetic traces that follow a consistent, domain-specific schema. 
This SFT phase establishes a reasoning prior, biasing the model toward producing traces that expose individual claims in predictable positions for verification.
This enables the downstream verifier to automatically extract those claims reliably using syntactic pattern matching rather than semantic parsing.

Concretely, the synthetic traces are
formatted according to a domain-specific schema that makes each intermediate reasoning step explicit.
The schema specifies what claims the model should make, in what order, and in what format, effectively defining a valid reasoning trace for the task.
The model is fine-tuned on these traces using standard cross-entropy loss prior to RL training.
The main goal of the SFT phase is to teach the model how to express its reasoning in a verifiable form, while the correctness of the expressed claims is shaped during the RL phase through the reasoning reward (see Figure~\ref{fig:sample_trace} for an example in chess).

\subsection{Reasoning verification and reward}
\label{ss:rv-reward}

Given a structured trace $\tau$, the reasoning verifier $\mathcal{V}$ extracts a set of claims $\{ c_1, c_2, \dots, c_K \}$ corresponding to predefined subtasks using deterministic parsing rules.
Each claim $c_k$ is evaluated against ground-truth signals available in the domain (e.g., rules, engine outputs, or consistency checks) to produce a subtask reward $r_k \in [0, 1]$.
Subtasks for which a claim is not applicable are excluded from reward aggregation for that instance.
The reasoning reward is an aggregation of subtask rewards:
\begin{equation*}
    R_{\text{reason}}(x, \tau) = \sum_{k=1}^{K} w_k \cdot r_k \quad \text{s.t.} \quad \sum_{k=1}^{K} w_k = 1,\; w_k \ge 0 \;\; \forall k \in \{1, \ldots, K\},
\end{equation*} 
where $w_k$ are aggregation weights.
In the simplest case, these weights are fixed and uniform; in our adaptive scheme they are dynamically adjusted (see Section~\ref{ss:adaptive-weighting}).

A key advantage of this approach is that no LLM judge is required at training time; 
claim extraction is automated thanks to the structured format enforced by the SFT prior.  
Verification of each claim is likewise cheap.
It reduces to numerical comparison, rule lookup, or deterministic solver queries depending on the subtask.

\subsection{Adaptive reward weighting}
\label{ss:adaptive-weighting}

A fixed aggregation of rewards is suboptimal when the model's competence across subtasks is highly uneven.
To address this, we dynamically adjust $w_k$ based on a running estimate of model performance, allocating more weight to subtasks with greater room for improvement.

\paragraph{EMA-based performance tracking.}
After each training step $t$, we compute the mean subtask reward $\bar{r}_{k}^{(t)}$ across the batch for each subtask $k$.
We maintain an exponential moving average (EMA) $\mu_k^{(t)}$ of the batch means:
\begin{equation*}
    \mu_k^{(t)} \;=\; (1 - \alpha) \cdot \mu_k^{(t-1)} + \alpha \cdot \bar{r}_{k}^{(t)},
\end{equation*} 
where $\alpha \in (0, 1)$ is the decay rate.
This provides a running estimate of the model's performance on subtask $k$ that adapts over the course of training while smoothing the inherent noise in batch-level statistics.
We initialize $\mu_k^{(0)} = 0.5$ for all $k$ (an uninformative prior) or set it based on prior knowledge of subtask difficulty.

\paragraph{Adaptive weighting.}

We define the \textit{headroom} of subtask $k$ at step $t$ as $h_{k}^{(t)} = 1 - \mu_{k}^{(t)}$, which measures the remaining room for improvement.
A subtask the model has largely mastered has low headroom, while a persistently difficult subtask retains high headroom.
Aggregation weights are computed via a temperature-scaled softmax over headrooms:
\begin{equation*}
    \tilde{w}_{k}^{(t)} \;=\;
    \frac{\exp(h_{k}^{(t)} / T)}{\sum_{j=1}^{K} \exp(h_{j}^{(t)} / T)},
\end{equation*}
where $T$ is the temperature controlling the sharpness of the weighting.
A minimum weight floor is then applied, and the weights are renormalized to obtain $w_{k}^{(t)}$.
The floor ensures that no subtask is entirely neglected even when the model performs well on it, preserving training signal across all components.
The final reward at step $t$ is then:
\begin{equation*}
    R(x, \tau, \hat{y}) \;=\;  R_{\text{form}}(\hat{y}) + R_{\text{acc}}(\hat{y}, y^{*}) + \lambda \sum_{k=1}^{K} w_{k}^{(t)} \cdot r_{k},
\end{equation*}
where the weights $w_{k}^{(t)}$ are updated at each step.
This mechanism implicitly induces a dynamic curriculum over reasoning subtasks, directing learning toward components where the model has the greatest potential for improvement (see Figure~\ref{fig:adapt-demo} for an illustration).

\section{Experiments}
\label{s:experiments}

\subsection{Setup}

\paragraph{Task.}
For empirical evaluation, we consider the task of predicting optimal chess moves.
Chess provides clear rules, reliable engine signals, and naturally decomposable subtasks, making it an ideal testbed for studying both prediction accuracy and reasoning quality.
Given a position encoded in Forsyth-Edwards Notation (FEN) (e.g., Figure~\ref{fig:trace_comp}), the model generates a trace analyzing candidate moves within \texttt{<think>} tags, followed by the final move prediction in \texttt{<answer>} tags.
See Appendix~\ref{app:rv_details} for background on chess concepts.

\paragraph{Evaluation.}
We evaluate both prediction accuracy and reasoning quality on held-out data.
Prediction accuracy is measured as the fraction of positions for which the model correctly predicts the optimal move, with Elo rating reported as a complementary measure of playing strength.
We additionally report candidate coverage, defined as the fraction of the engine's top moves included among the model's candidate moves.
Reasoning quality is assessed along three dimensions, namely evaluation accuracy, continuation analysis, and internal consistency (see Appendix~\ref{app:rv_eval_details} for additional metrics and results):
\begin{itemize}[leftmargin=7mm] %
    \item \textbf{Win-rate MAE (eval accuracy):} mean absolute error between the model's claimed win rates and engine's, where lower values indicate more accurate positional evaluation.
    \item \textbf{PV overlap (continuation analysis):} fraction of predicted continuation moves matching the engine principal variation, averaged across depth.
    \item \textbf{Logic consistency (internal consistency):} fraction of positions where the concluded best move is the candidate with the highest win rate, measuring self-consistency.
\end{itemize}

\paragraph{Data.}

\begin{wrapfigure}{r}{0.35\linewidth}
    \centering
    \vspace{-12pt}
    \includegraphics[width=1.0\linewidth]{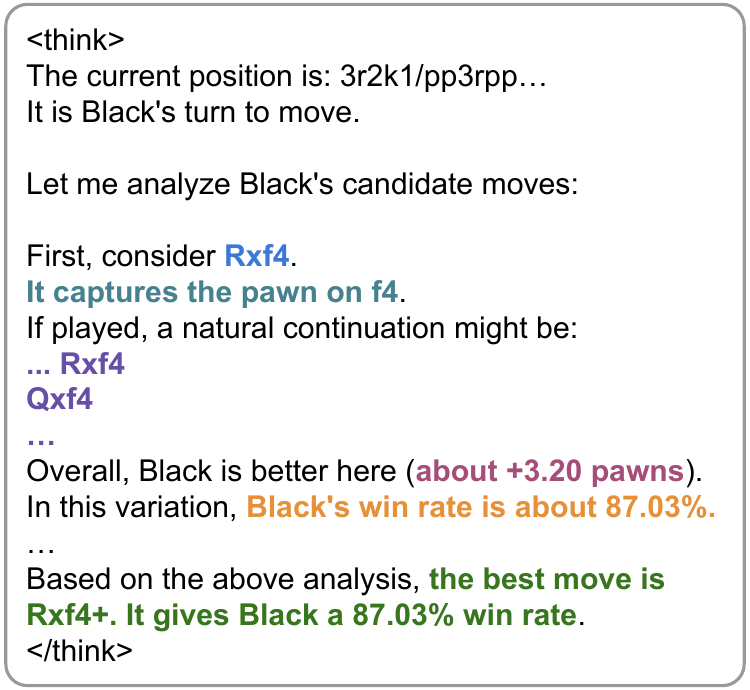}
    \vspace{-15pt}
    \caption{\textbf{Example synthetic reasoning trace.} Colors indicate individual verifiable claims.}
    \vspace{-10pt}
    \label{fig:sample_trace}
\end{wrapfigure}

For SFT, we construct synthetic traces from the Lichess Evaluations Database consisting of a large collection of chess positions analyzed by Stockfish, a widely used open-source chess engine.
Each entry provides a position in FEN, a list of moves with centipawn scores (a numerical measure of positional advantage), mate evaluations if applicable, and principal variation (PV) representing the engine's predicted best continuation from the position.
Unlike rule-based signals such as captures, PV and win-rate estimation require reasoning about long-term positional consequences and are therefore considerably more challenging for general language models.
For each position, we construct a trace analyzing the top-3 moves with respect to captures, checks, mates, positional advantage (pawn scores), win rate, and PV (Figure~\ref{fig:sample_trace}).
We sample 10k positions for SFT, which we find sufficient to reliably instill the structured reasoning prior.
In our ablations, we also include models trained on 1M positions to assess the effect of stronger pre-RL domain knowledge.

For RL training and evaluation, we use the Lichess Puzzle Database, which contains tactically sharp positions where deviating from the optimal move sequence substantially worsens the position.
We sample 50k, 1k, and 1k puzzles for train, validation, and test splits, respectively.
Each puzzle provides a move sequence from an initial FEN, which is flattened into multiple training examples following prior work~\citep{ruoss2024amortized}.
Because the puzzle database does not include Stockfish analysis, we run the engine independently on all samples: training positions are analyzed with search depth 25, while held-out positions use depth 30 to obtain higher-precision evaluation signals.

\paragraph{Training.}

We use Qwen3-8B~\citep{yang2025qwen3} and DeepSeek-R1-Distill-Llama-8B~\citep{guo2025deepseek} in our main experiments.\footnote{The code and implementation details are available at \url{https://github.com/kykim0/VPS}.}
Training proceeds in two stages: SFT to instill a structured reasoning prior, followed by GRPO initialized from the SFT checkpoint.
The accuracy reward is computed using a pre-trained action-value network $Q_\theta(s, a)$, a 270M-parameter 16-layer decoder-only Transformer from~\cite{ruoss2024amortized}, trained to predict the post-move win probability.
During RL, the network is queried with the model's predicted move $\hat{a}$ at each position $s$ to compute $R_{\text{acc}} = Q_\theta(s, \hat{a}) \in [0, 1]$.
This dense reward has been shown to improve convergence compared to binary correctness signals~\citep{hwang2025can}.

The reasoning reward is computed by a verifier that evaluates chess-specific subtasks.
Each subtask reward $r_k$ evaluates a claim extracted from the model's trace against the corresponding ground truth.
For numerical subtasks, we define a tolerance interval within which the model receives full credit.
Outside this interval, rewards decay linearly until reaching zero at a threshold.
This provides a smooth learning signal while remaining robust to minor engine inconsistencies.
For binary subtasks (e.g., capture, check), rewards are computed using exact correctness.
See Appendix~\ref{app:reward_details} for details on the reward design.

\subsection{Main results}

\paragraph{Effectiveness of adaptive reasoning supervision.}

\begin{table}[t]
    \centering
    \footnotesize  %
    \caption{\textbf{Accuracy and reasoning quality comparison.} Prediction accuracy and reasoning quality metrics
             for Qwen3-8B and DeepSeek-R1-Distill-Llama-8B. Reasoning quality is evaluated using a deterministic verifier on held-out positions; lower win-rate MAE and higher values on other metrics indicate better reasoning.}
    \label{table:main-table}
    \vspace{-5pt}
    \begin{tabular}{l*{7}{c}}
        \toprule[1pt]
        \multirow{2.5}{*}{\textbf{Model}} & \multicolumn{3}{c}{\textbf{{Prediction Accuracy}}} & \multicolumn{3}{c}{\textbf{{Reasoning Quality}}} \\
        \cmidrule(lr){2-4} \cmidrule(lr){5-7}
         & Top-1 Acc. $\uparrow$ & Elo $\uparrow$ & Coverage $\uparrow$ & WR MAE $\downarrow$ & PV Overlap $\uparrow$ & Consist. $\uparrow$ \\
        \midrule[0.75pt]
        \multicolumn{6}{l}{\textit{Qwen3-8B}} \\
        \hspace{0.3em} SFT only & 0.212 & 1160 & 0.207 & \underline{0.310} & \underline{0.323} & \underline{0.961} \\
        \hspace{0.3em} SFT + GRPO & \underline{0.534} & \underline{1670} & \underline{0.302} & 0.452 & 0.183 & 0.721 \\
        \hspace{0.3em}  {\mname} & \textbf{0.536} & \textbf{1706} & \textbf{0.441} & \textbf{0.218} & \textbf{0.376} & \textbf{0.978} \\
        \midrule[0.50pt]
        \multicolumn{6}{l}{\textit{R1-Distill-Llama-8B}} \\
        \hspace{0.3em} SFT only & 0.215 & 996 & 0.226 & \underline{0.346} & \underline{0.341} & \underline{0.978} \\
        \hspace{0.3em} SFT + GRPO & \underline{0.541} & \underline{1670} & \underline{0.385} & 0.732 & 0.330 & 0.312 \\
        \hspace{0.3em} {\mname} & \textbf{0.545} & \textbf{1744} & \textbf{0.508} & \textbf{0.286} & \textbf{0.362} & \textbf{0.985} \\
        \bottomrule[1pt]
    \end{tabular}
    \vspace{-5pt}
\end{table}

Optimizing for accuracy alone yields strong performance but severely degrades reasoning quality.
Table~\ref{table:main-table} compares SFT, SFT followed by outcome-only GRPO, and {\mname}, illustrating the trade-off between task performance and reasoning quality.
Note that top-1 accuracy in the range of 0.53--0.55 is non-trivial: positions contain 27 legal moves on average, and the task requires identifying the single optimal move rather than merely a reasonable one.

SFT alone achieves modest accuracy (0.212--0.215) with well-structured, internally consistent reasoning.
Outcome-only GRPO improves accuracy (0.534--0.541), but sharply degrades reasoning quality.
Win-rate MAE increases by 46\% and 112\% relative to SFT, while internal consistency drops to 0.721 and 0.312.
For R1-Distill-Llama-8B, this implies that nearly 70\% of predictions contradict the model's own win rate analysis, indicating actively incoherent reasoning under outcome-only optimization.
Candidate diversity also reveals reward gaming: the number of unique candidates analyzed for Qwen3-8B drops from 2.20 to 1.65, suggesting repetition of top moves rather than genuine comparison.

In contrast, {\mname} matches the accuracy of outcome-only GRPO while substantially improving reasoning quality.
Win-rate MAE decreases to 0.218 and 0.286, and internal consistency reaches near-perfect levels (0.978 and 0.985).
Moreover, coverage more than doubles relative to SFT for both models, with the average unique candidates for Qwen3-8B approaching 3.0, indicating more thorough analysis of engine-aligned moves.
PV overlap shows more modest gains, as predicting the optimal continuation requires deep search, even for strong engines (see Section~\ref{ss:ablations} for a related discussion).
Overall, {\mname} achieves a substantially better trade-off between prediction accuracy and reasoning quality across both models.
See Appendix~\ref{app:more_reason_metrics} for further results.

\paragraph{Accuracy-controlled reasoning comparison.}

\begin{table}[t]
    \centering
    \footnotesize
    \caption{\textbf{Accuracy-controlled reasoning comparison.}
    Reasoning quality is evaluated using three LLM judges (GPT-4o, Claude Opus 4.6, Gemini 3 Flash) along four dimensions (1--5 scale).
    We compare the highest-accuracy GRPO-only checkpoint with a {\mname} checkpoint at matched accuracy, and include the GRPO-only checkpoint at the same training step.}
    \label{table:llm-judge}
    \vspace{-5pt}
    \begin{tabular}{lc*{5}{c}}
        \toprule[1pt]
        \textbf{Model}
            & Acc. $\uparrow$
            & Relev. $\uparrow$ 
            & Compl. $\uparrow$ 
            & Clarity $\uparrow$ 
            & Fluency $\uparrow$
            & Overall $\uparrow$ \\
        \midrule[0.75pt]
        \multicolumn{6}{l}{\textit{Qwen3-8B}} \\
        \hspace{0.5em} GRPO only \scriptsize{(best acc.)}    & 0.507 & \underline{2.501} & \underline{2.296} & \underline{2.429} & 3.102 & 2.582 \\
        \hspace{0.5em} {\mname} \scriptsize{(matched acc.)}  & 0.505 & \textbf{2.856} & \textbf{3.239} & \textbf{3.751} & \textbf{3.833} & \textbf{3.420} \\
        \hspace{0.5em} GRPO only \scriptsize{(same step)}    & 0.500 & 2.484 & 2.292 & 2.390 & \underline{3.275} & \underline{2.610} \\
        \midrule[0.50pt]
        \multicolumn{6}{l}{\textit{R1-Distill-Llama-8B}} \\
        \hspace{0.5em} GRPO only \scriptsize{(best acc.)}    & 0.539 & 2.057 & 1.862 & 1.985 & 3.164 & 2.267 \\
        \hspace{0.5em} {\mname} \scriptsize{(matched acc.)}  & 0.539 & \textbf{2.648} & \textbf{3.322} & \textbf{3.386} & \textbf{3.729} & \textbf{3.271} \\
        \hspace{0.5em} GRPO only \scriptsize{(same step)}    & 0.532 & \underline{2.223} & \underline{2.052} & \underline{2.108} & \underline{3.242} & \underline{2.406} \\
        \bottomrule[1pt]
    \end{tabular}
    \vspace{-5pt}
\end{table}

While Table~\ref{table:main-table} isolates the effect of adding process supervision on top of a shared SFT prior, we ask a complementary question: can outcome-only GRPO learn sound reasoning without a structured prior?
To assess this fairly, we compare outcome-only GRPO against {\mname}, controlling for both accuracy and training step.
We select the highest accuracy GRPO-only checkpoint, a {\mname} checkpoint at comparable accuracy (typically earlier), and include the GRPO-only checkpoint at the same step.
Because outcome-only GRPO produces free-form reasoning, we adopt the LLM chess commentary evaluation framework of~\citet{kim2025bridging}, shown to align with human judgments.
We evaluate relevance, completeness, clarity, and fluency on a 1--5 scale using three judge models, reporting the mean and standard deviation.

Table~\ref{table:llm-judge} reports reasoning quality scores averaged across three LLM judges (GPT-4o, Claude Opus 4.6, Gemini 3 Flash).
Across both models, {\mname} consistently produces higher-quality reasoning than GRPO-only at matched accuracy, with large margins in completeness, clarity, and fluency.
Notably, the GRPO-only best-accuracy checkpoint, despite achieving higher accuracy, is often judged to produce worse reasoning than the earlier checkpoint.
This pattern is consistent with the shortcut-seeking behavior described in Section~\ref{ss:ablations}, where extended outcome-only training degrades reasoning quality even as accuracy plateaus or improves marginally.
Relevance scores are generally lower across methods, as traces must analyze candidates aligned with the engine's top moves using position-specific reasoning, which is more challenging than analyzing arbitrary legal moves.
Overall, even at matched accuracy, structured reasoning with adaptive supervision achieves substantially higher reasoning quality.
See Appendix~\ref{app:judge-eval} for detailed analysis and further results.

\subsection{Generalization to math reasoning}
\label{ss:gsm8k-results}

\begin{wraptable}{r}{0.50\textwidth}
    \vspace{-12pt}
    \centering
    \footnotesize  %
    \caption{\textbf{GSM8K results.} Outcome-only RL improves accuracy but degrades reasoning quality, whereas {\mname} preserves both.}
    \label{table:gsm8k-table}
    \vspace{-5pt}
    \begin{tabular}{lccc}
    \toprule[1pt]
    \textbf{Method} & Acc. & Step Arith. & Consist. \\
    \midrule[0.75pt]
    \text{SFT only}   & $0.446$ & $0.638 $ & $0.821$ \\
    \text{GRPO-only}  & $0.760$ & $0.681 $ & $0.647$ \\
    \text{SFT + GRPO} & $0.760$ & $0.658 $ & $0.621$ \\
    \text{{\mname}}   & $\bf{0.763}$ & $\bf{0.905}$ & $\bf{0.889}$ \\
    \bottomrule[1pt]
    \end{tabular}
    \vspace{-5pt}
\end{wraptable}

To assess whether our findings generalize beyond chess, we additionally evaluate {\mname} on GSM8K~\citep{cobbe2021training}.
We use Qwen3-0.6B and follow the same training procedure as in the chess experiments.
Outcome-only RL solely optimizes final-answer correctness, whereas {\mname} additionally supervises intermediate arithmetic claims and consistency.
We evaluate accuracy together with step arithmetic (the fraction of arithmetic claims that evaluate to the stated value) and answer consistency (whether the final answer matches the conclusion of the reasoning trace).
The key findings are consistent with the chess experiments.
Outcome-only GRPO substantially improves final-answer accuracy over the SFT baseline but markedly degrades answer consistency, with only modest gains in step arithmetic.
GRPO-only exhibits a similar trade-off, achieving competitive accuracy with slightly better arithmetic correctness but lower consistency.
In contrast, {\mname} achieves comparable final-answer accuracy to outcome-only GRPO while substantially improving reasoning quality.
It markedly improves both step arithmetic and answer consistency, surpassing the SFT baseline on both reasoning metrics.
Overall, these findings mirror the chess experiments, suggesting that outcome-only RL can improve final-answer accuracy at the expense of reasoning quality, whereas verifiable process supervision largely mitigates this trade-off across both chess and mathematical reasoning.

\subsection{Ablations and analysis}
\label{ss:ablations}

\paragraph{Effect of adaptive weighting and SFT scale.}

\begin{figure}[t]
     \centering
     \includegraphics[width=\linewidth]{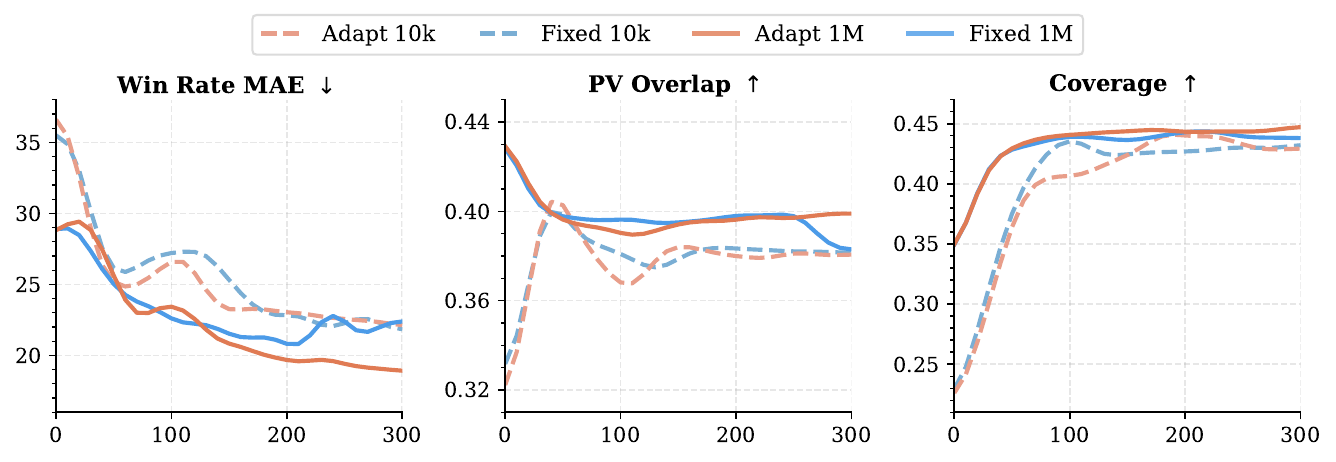}
     \vspace{-15pt}
     \caption{\textbf{Reward weighting across SFT scales.} Adaptive weighting improves the hardest reasoning subtasks, particularly when models start with stronger domain knowledge.
     }
     \label{fig:weight-comp}
     \vspace{-5pt}
\end{figure}

Figure~\ref{fig:weight-comp} compares adaptive and fixed weighting for models initialized from SFT trained on 10k and 1M positions.
Adaptive weighting consistently improves the hardest subtasks, win-rate estimation and PV prediction, while maintaining competitive accuracy.
Gains are modest in the 10k SFT setting, likely because (i) the subtask rewards are already designed to induce a curriculum, and (ii) reweighting provides limited benefit when models enter RL without sufficient domain knowledge.
In contrast, benefits are substantially larger in the 1M SFT setting, where models begin with stronger domain competence.
Adaptive weighting reduces win-rate MAE by 15.1\% relative to fixed weighting (18.66 vs.~21.97) and improves PV overlap ($0.38 \rightarrow 0.40$).
PV overlap initially decreases under both schemes, which we attribute to RL pressure shifting toward accurate move prediction at the expense of deeper continuation quality in 1M SFT models; it recovers under adaptive weighting but persists under fixed weighting.
Similarly, win-rate MAE plateaus under fixed weighting but consistently improves with adaptive weighting.
This highlights the trade-off in fixed weighting: after easy gains are exhausted, difficult subtasks are deprioritized, while adaptive weighting shifts focus to remaining errors.

\paragraph{Qualitative comparison of reasoning quality.}

\begin{figure}[t]
     \centering
     \begin{subfigure}[b]{0.32\textwidth}
         \centering
         \includegraphics[width=\textwidth]{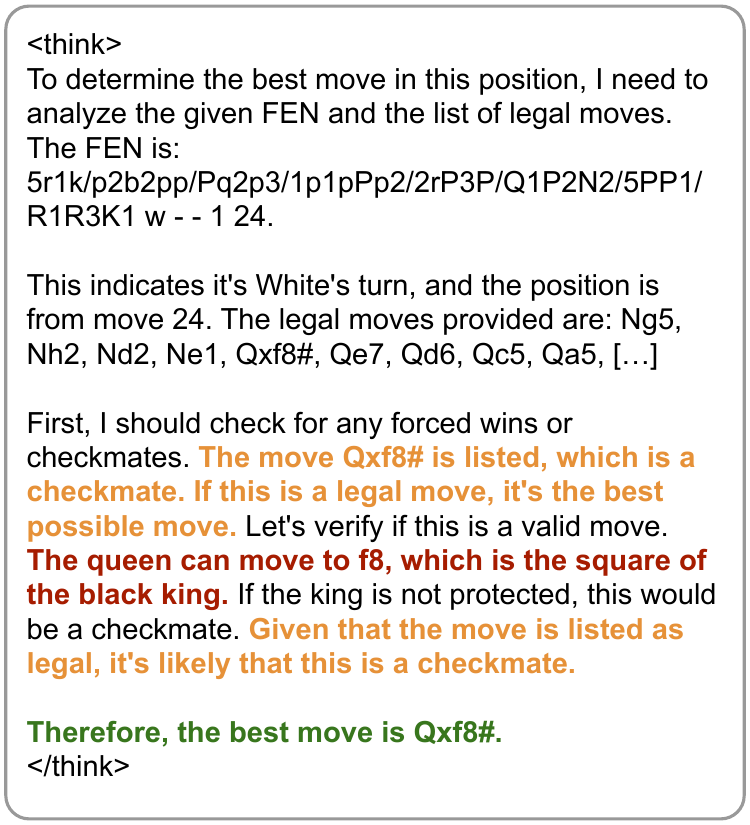}
         \caption{GRPO only}
     \end{subfigure}
     \hfill
     \begin{subfigure}[b]{0.32\textwidth}
         \centering
         \includegraphics[width=\textwidth]{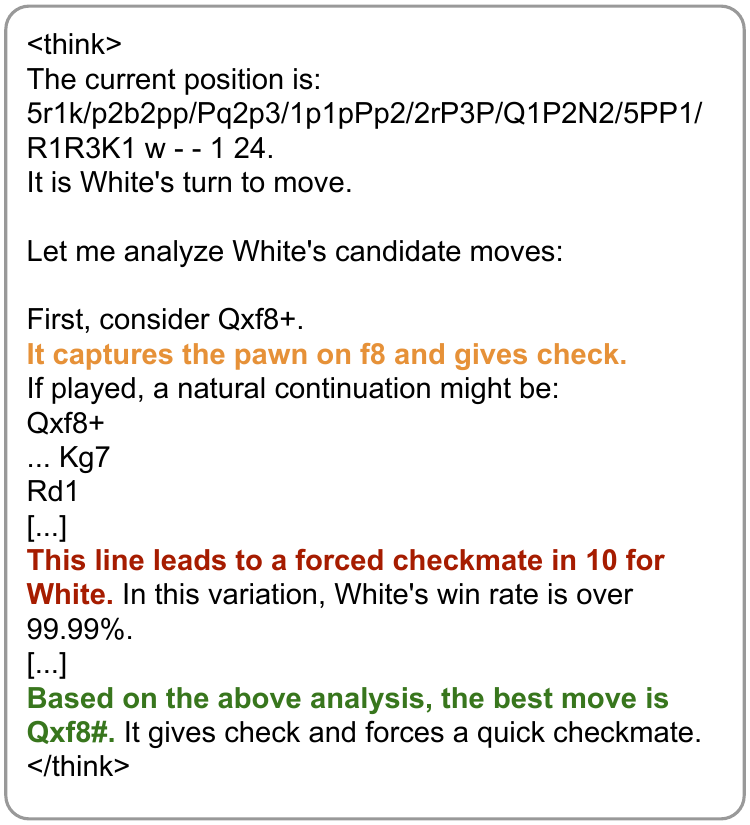}
         \caption{SFT + GRPO}
     \end{subfigure}
     \hfill
     \begin{subfigure}[b]{0.32\textwidth}
         \centering
         \includegraphics[width=\textwidth]{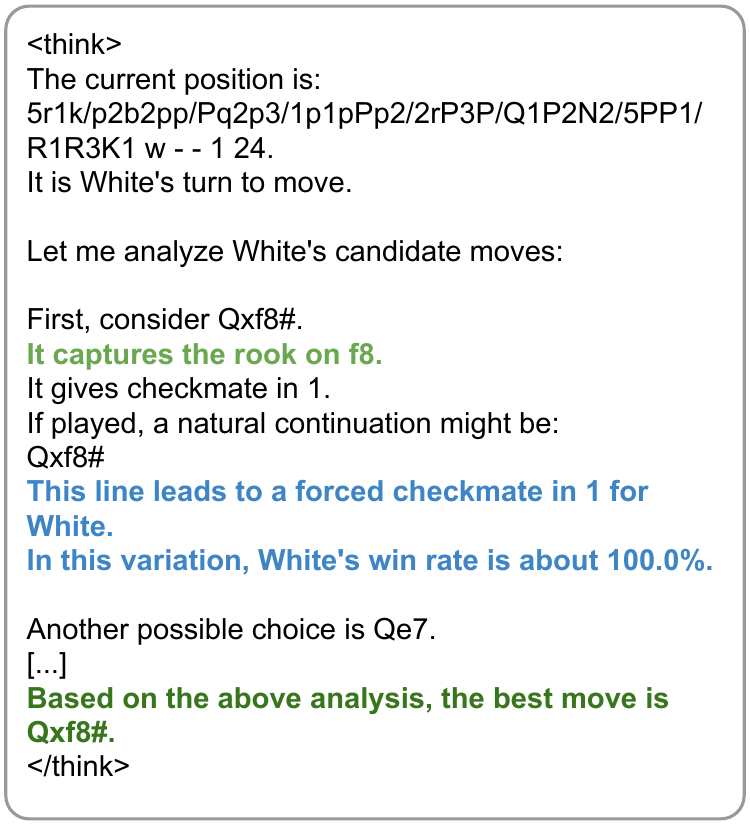}
         \caption{{\mname}}
     \end{subfigure}
     \vspace{-5pt}
     \caption{\textbf{Qualitative reasoning comparison.} Outcome-only optimization produces flawed or inconsistent reasoning, while {\mname} yields concise and factually grounded explanations.}
     \label{fig:trace_comp}
     \vspace{-5pt}
\end{figure}

We compare reasoning traces from outcome-only GRPO, SFT-initialized GRPO, and {\mname} (Figure~\ref{fig:trace_comp}), illustrating substantial differences in reasoning quality despite identical answers.
Outcome-only GRPO exhibits several failure modes, including redundant enumeration of legal moves, reliance on superficial notational cues (e.g., ``\texttt{\#}'' indicating checkmate), and inconsistent explanations about the board state.
SFT initialization improves fluency but still leaves factual errors and shows signs of reward hacking: it conflates notation (e.g., \texttt{Qxf8+} vs.~\texttt{Qxf8\#}), misidentifies captures, and repeats claims such as ``forced checkmate in 10.'' 
In contrast, {\mname} produces concise, accurate, and internally consistent reasoning, correctly identifying the captured piece, the mate-in-one outcome, and a coherent PV. %
These examples show that outcome-only optimization can yield flawed reasoning despite correct answers, whereas verifiable process supervision enforces correctness and consistency at the level of individual reasoning steps.

\paragraph{Effect of reasoning budget on convergence and quality.}

\begin{wrapfigure}{r}{0.32\linewidth}
    \vspace{-15pt}
    \centering
    \includegraphics[width=0.95\linewidth]{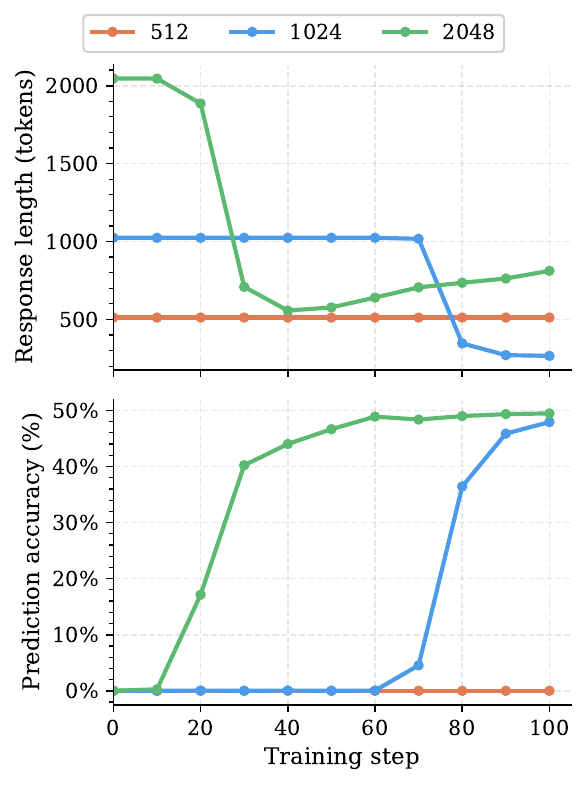}
    \vspace{-8pt}
    \caption{\textbf{Reasoning space.} Outcome-only RL produces variable reasoning patterns without a structured prior. 
    }
    \vspace{-10pt}
    \label{fig:reason_space}
\end{wrapfigure}

We study how the available reasoning budget shapes the strategies learned under outcome-only RL.
In these experiments, we train Qwen3-4B with outcome-only rewards while varying the maximum output length.
Without a structured prior, models converge to different shortcut strategies depending on their reasoning budget, despite reaching similar accuracy (Figure~\ref{fig:reason_space}).
With a 2048-token limit, models often produce long exploratory traces that iterate over legal moves without meaningful analysis.
Accuracy varies sharply with trace length: 256--512 token traces achieve 86\% accuracy, whereas those exceeding 1024 tokens drop to 27\%, suggesting that longer traces reflect confusion rather than deeper reasoning.
With a 1024-token limit, models are forced into more concise reasoning patterns, converging to shorter traces ($\sim$200 tokens) at comparable accuracy.
In contrast, a 512-token limit prevents convergence, suggesting a lower bound on the reasoning space needed to discover successful paths.
Overall, these results show that outcome-only RL produces shortcut strategies whose form depends more on the allowed reasoning space than on principled multi-step reasoning.
See Appendix~\ref{app:reason_budget} for further analysis results.

\section{Conclusion}
\label{s:conclusion}

In this work, we introduce verifiable process supervision ({\mname}), a post-training framework that jointly optimizes accuracy and reasoning quality via deterministic supervision of intermediate claims.
By combining structured reasoning induced through supervised fine-tuning with deterministic verification, {\mname} enables scalable process supervision without learned reward models or LLM judges.
We further introduce adaptive weighting that prioritizes components with the greatest remaining headroom, creating a curriculum over reasoning subskills that improves training efficiency.
Across chess and math reasoning, we show that outcome-only RL can improve task performance while degrading reasoning quality, whereas {\mname} largely mitigates this trade-off while preserving accuracy.
These findings suggest that deterministic verification of intermediate reasoning provides a practical foundation for training language models whose reasoning is accurate, consistent, and reliable.

\section*{Acknowledgments}

This research was supported in part by the Institute for Information \& Communications Technology Planning \& Evaluation (IITP) grant funded by the Korea government (MSIT) (RS-2019-II190075, Artificial Intelligence Graduate School Support Program (KAIST); RS-2022-II220953, Self-directed AI Agents with Problem-solving Capability; RS-2022-II220959, Few-shot Learning of Causal Inference in Vision and Language for Decision Making).
The authors thank Sentient Labs for generously providing compute resources for this work.

\bibliography{colm2026_conference}
\bibliographystyle{colm2026_conference}

\newpage
\appendix
\section{Experimental details}
\label{app:exp_details}

\subsection{Reasoning verification}
\label{app:rv_details}

\paragraph{Chess concepts.}

We briefly review key chess concepts.
Board positions are represented using Forsyth-Edwards Notation (FEN), which encodes piece placement, active color, castling rights, en passant targets, and move counters.
Moves are represented in Standard Algebraic Notation (SAN), which specifies the moving piece, destination square, and optional annotations (e.g., captures, checks, and checkmates).

The principal variation (PV) is the sequence of moves the engine evaluates as the best continuation from the current position.
The centipawn score (cp) is a numerical evaluation of a position measured in hundredths of a pawn, where positive values favor the side to move.
The win rate is derived from the centipawn score via the standard Lichess formula\footnote{\url{https://lichess.org/page/accuracy}}:
\begin{equation*}
    \text{Win-Rate} = 50 + 50 \left( \frac{2}{1 + e^{-0.00368208 \cdot \text{cp}}} - 1 \right),
\end{equation*}
which estimates the probability of winning from a given position.
Capture, check, and mate refer respectively to whether a move takes an opponent's piece, puts the opponent's king in check, or delivers checkmate.

Concepts such as PV, centipawn score, and win rate are predictions produced by a strong chess engine and require evaluating the long-term consequences of moves.
As a result, these are difficult for general language models to estimate accurately without exposure to sufficient engine-generated data.
In contrast, capture and check can be determined directly from the rules of chess and are therefore easier for language models that have learned the game mechanics.

\paragraph{Reward design.}
\label{app:reward_details}

We provide additional details on the design of subtask rewards used in our experiments.
For numerical subtasks, we define a \textit{flat zone} $[0,\delta_k]$ within which the model receives full credit.
The flat zone accounts for small fluctuations in engine evaluations, ensuring that near-identical positions are not penalized due to minor evaluation noise.
Beyond this, the reward decays linearly until reaching zero at threshold $\Delta_k$:
\begin{equation*}
    r_k(d_k) =
    \begin{cases}
    1, & d_k \le \delta_k, \\
    1 - \frac{d_k - \delta_k}{\Delta_k - \delta_k}, & \delta_k < d_k \le \Delta_k, \\
    0, & d_k > \Delta_k,
    \end{cases}
\end{equation*}
where $d_k$ denotes the absolute difference between the claimed and ground-truth values for subtask $k$.
This design provides a smooth learning signal: near-correct predictions receive full reward, while progressively larger errors receive proportionally smaller rewards.

Below we provide definitions of the subtask rewards used in our experiments.
All rewards are computed from claims explicitly stated in the model's reasoning trace.
If a required claim is missing, the corresponding subtask receives zero reward.
\begin{itemize}[leftmargin=7mm]
    \item \textbf{Win rate:} $d = |\text{claimed win rate} - \text{engine win rate}|$, with flat zone $\delta = 0.1$ and threshold $\Delta = 10.0$ (in percentage points).
    \item \textbf{Pawn score:} $d = |\text{claimed pawn score} - \text{engine pawn score}|$, with flat zone $\delta = 0.1$ and threshold $\Delta = 3.0$. Sign mismatches (i.e., claiming an advantage when Stockfish finds a disadvantage) receive zero reward. When both claimed and ground-truth evaluations exceed 5 pawns in magnitude, the effective difference is capped at 2.0 pawns, as precise discrimination is less meaningful in clearly winning or losing positions.
    \item \textbf{Principal variation:} The first move of the claimed PV must match Stockfish's PV since this is the move being analyzed. For subsequent moves, we compute a depth-weighted prefix agreement with weights $[0.4, 0.3, 0.2, 0.1]$ for depths 1 through 4, assigning higher reward when the model's continuation matches the engine's PV at earlier depths.
    \item \textbf{Capture:} Binary reward for correctly identifying whether the move is a capture and, if so, naming the correct piece type and target square. Missing a capture or claiming a capture when none occurs both receive zero reward.
    \item \textbf{Check:} Binary reward for correctly identifying whether the move delivers check.
    \item \textbf{Mate:} Reward for correctly identifying a forced checkmate, with partial credit for near-correct mate-in-$n$ predictions.
    \item \textbf{Logic consistency:} Checks whether the model's final move selection corresponds to the move with the highest predicted win rate among the moves it analyzed. This reward encourages internal consistency by checking whether the model's final move selection corresponds to the analyzed move with the highest predicted win rate. It is independent of whether the analyzed moves match Stockfish's top candidates.
\end{itemize}

\paragraph{Temperature adaptation.}

The fixed temperature $T$ used in the softmax weighting applies the same degree of weight concentration regardless of how uneven the model's subtask performance is.
When headrooms are similar across subtasks, strong concentration is unnecessary, whereas larger disparities may benefit from sharper weighting.

As a minor extension, we also consider an adaptive variant where the temperature depends on the spread of current headrooms.
Specifically, we scale the base temperature inversely with the standard deviation of the headroom vector:
\begin{equation*}
    T^{(t)} = \frac{T_0}{1 + \sigma^{(t)}},
\end{equation*}
where $T_0$ is the base temperature and $\sigma^{(t)} = \text{std}(h_{1}^{(t)}, \dots, h_{K}^{(t)})$ is the standard deviation of headrooms at step $t$.
When headrooms are nearly uniform ($\sigma^{(t)} \approx 0$), $T^{(t)} \approx T_0$ and weighting remains relatively diffuse.
When headroom varies more strongly, $T^{(t)}$ decreases, sharpening the softmax and concentrating weight on the most underperforming subtasks.
In practice, a fixed temperature (e.g., $T=1.0$) often works well, although the adaptive variant can reduce the need for manual tuning.

\subsection{Training details}

\paragraph{Hyperparameters.}

\begin{table}[!t]
    \centering
    \caption{\textbf{Supervised fine-tuning hyperparameters.}}
    \label{tab:hparams-sft}
    \vspace{-5pt}
    \setlength{\tabcolsep}{8pt}
    \begin{tabular}{lll}
    \toprule
     & \textbf{Hyperparameter} & \textbf{Value} \\
    \midrule[0.75pt]
    \textbf{Training} & Epochs & 2 \\
                      & Batch size & 256 \\
                      & Gradient clipping & 1.0 \\
                      & Precision & bf16 \\
    \midrule[0.50pt]
    \textbf{Optimization} & Learning rate & 1e-5 \\
                          & LR scheduler & cosine \\
                          & Warmup ratio & 0.1 \\
                          & Weight decay & 0.01 \\
    \bottomrule[1pt]
    \end{tabular}
    \vspace{-5pt}
\end{table}

\begin{table}[!t]
    \centering
    \caption{\textbf{Reinforcement learning hyperparameters.}}
    \label{tab:hparams-rl}
    \vspace{-5pt}
    \setlength{\tabcolsep}{8pt}
    \begin{tabular}{lll}
    \toprule[1pt]
     & \textbf{Hyperparameter} & \textbf{Value} \\
    \midrule[0.75pt]
    \textbf{Training} & Training steps & 300 \\
                      & Learning rate & 1e-6 \\
                      & PPO epochs per rollout & 1 \\
                      & KL coefficient & 0.001 \\
                      & Discount factor ($\gamma$) & 1.0 \\
    \midrule[0.50pt]
    \textbf{Data} & Prompts per batch & 128 \\
                  & Mini-batch size & 128 \\
    \midrule[0.50pt]
    \textbf{Rollout} & Samples per prompt ($n$) & 8 \\
                     & Temperature & 1.0 \\
    \midrule[0.50pt]
    \textbf{{\mname}} & Reasoning reward weight ($\lambda$) & 1.0 \\
                      & Effective EMA smoothing ($\alpha$) & 0.128 \\
                      & Adaptive weighting temperature & 1.0 \\
    \bottomrule[1pt]
    \end{tabular}
    \vspace{-5pt}
\end{table}

Tables~\ref{tab:hparams-sft} and~\ref{tab:hparams-rl} summarize the SFT and RL hyperparameters used in our experiments.
The {\mname} parameters apply when reasoning supervision is enabled.
We use consistent settings across methods unless otherwise specified.

\paragraph{Runtime overhead.}

The additional computational overhead of {\mname} arises from two sources: (1) syntactic parsing of the generated trace, and (2) verification of extracted intermediate claims against ground-truth signals.
Parsing incurs negligible overhead due to the fixed structured reasoning format induced during SFT.
The cost of verification depends on the type of signal: rule-based checks (e.g., move legality or captures) are inexpensive, whereas verification of engine-dependent quantities, such as win rates and principal variations, is dominated by the runtime of the external chess engine.
In our implementation, all ground-truth engine signals are pre-computed offline before RL training.
Consequently, online verification consists only of lightweight comparisons against cached targets rather than repeated engine evaluations.
Under this implementation, training Qwen3-8B on the chess reasoning task using 4$\times$H100 GPUs took approximately 610 seconds per optimization step for outcome-only GRPO and 630 seconds for {\mname}, corresponding to only a modest runtime overhead.

\section{Evaluation details}
\label{app:eval_details}

\subsection{Reasoning quality evaluation}
\label{app:rv_eval_details}

All reasoning quality metrics are computed over unique candidate moves: if the model analyzes the same move multiple times, only the first occurrence is retained.
This prevents reward-hacking behavior (e.g., repeatedly analyzing a single well-predicted move) from artificially inflating accuracy.
For numerical metrics, we apply a worst-case penalty when the model fails to produce a claim for a move for which engine ground truth is available.
Specifically, we assign a penalty of 100 percentage points for win rate MAE, 10 pawns for pawn score MAE, and 0.0 for PV overlap rate.
This ensures that omitted claims are penalized rather than implicitly rewarded through a reduced denominator.
For samples with missing or degenerate reasoning traces, all applicable metrics receive their respective worst-case penalties.
Certain metrics are computed only when applicable: capture accuracy is evaluated only for positions involving captures, check accuracy only for moves delivering check, and mate MAE only for positions where the engine identifies a forced mate.
All remaining metrics (e.g., coverage and consistency) are computed over the full set.

\subsection{LLM judge evaluation}
\label{app:judge-eval}

\paragraph{Evaluation dimensions.}
We evaluate each reasoning trace along four dimensions, each rated on a 1--5 scale, following the design of the original LLM-based chess commentary evaluation framework~\citep{kim2025bridging}:
\begin{itemize}[leftmargin=7mm]
    \item \textit{Relevance} measures whether the analyzed moves are reasonable candidates for the position and whether their analysis is grounded in position-specific evidence.
    The Stockfish engine summary provides a reference for meaningful candidates.
    Reasoning that identifies strong candidate moves and supports them with concrete position-specific analysis receives higher scores than those that focus on irrelevant moves or generic statements (e.g., ``this is a check'').
    \item \textit{Completeness} measures whether the reasoning follows through on its analysis.
    High-scoring traces consider multiple candidate moves and provide meaningful analysis for each to support a conclusion, while traces that introduce candidates without analyzing them, or focus on a single move, score lower.
    \item \textit{Clarity} measures whether the reasoning is specific, precise, and unambiguous.
    Concrete evidence, such as named moves, continuations, centipawn evaluations, and win rates, scores higher than vague assertions (e.g., ``this improves the position'').
    \item \textit{Fluency} measures language quality, including grammatical correctness, sentence structure, and coherent transitions between ideas.
    It is independent of chess correctness.
    Repetitive or circular reasoning that adds no new information lowers the score.
\end{itemize}

\paragraph{Prompt construction.}
We provide the LLM judge with engine analysis as contextual grounding, rather than evaluating reasoning quality in isolation.
Engine analysis is provided as a qualitative summary derived from Stockfish evaluations.
While prior work provides two evaluations (the position before and after the move), we supply richer context by including the top-5 engine moves, enabling more accurate assessment of candidate relevance and analysis completeness.
The summary expresses the analyzed move's quality relationally (e.g., ``similar to best move'' or ``significantly worse than best move (over a pawn)'') along with the candidate move names and whether any lead to forced checkmate.

\paragraph{Score computation.}
Following the original design, we compute scores as probability-weighted expectations over the judge model's output distribution~\citep{kim2025bridging}.
Specifically, for a given dimension, the judge is prompted to respond with an integer score from 1 to 5.
We extract the log-probabilities of all score tokens (digits 1 through 5) at the position of the numeric token in the response and convert them into probabilities to compute the score:
\begin{equation*}
    \text{score}(\tau) = \sum_{s\in\{1, 2, 3, 4, 5\}} s \cdot p(s \mid \tau).
\end{equation*}
This produces a continuous score in $[1, 5]$ that captures the judge's uncertainty over adjacent ratings.
When log-probabilities are unavailable, we fall back to the parsed integer score.

\paragraph{Evaluation by individual judges.}

\begin{table}[t]
    \centering
    \footnotesize
    \caption{\textbf{Reasoning comparison with GPT-4o.} Accuracy-controlled evaluation of reasoning quality using GPT-4o as the judge.}
    \label{table:llm-judge-4o}
    \vspace{-5pt}
    \begin{tabular}{lc*{4}{c}}
        \toprule[1pt]
        \textbf{Model}
            & Relev. $\uparrow$ 
            & Compl. $\uparrow$ 
            & Clarity $\uparrow$ 
            & Fluency $\uparrow$
            & Overall $\uparrow$ \\
        \midrule[0.75pt]
        \multicolumn{6}{l}{\textit{Qwen3-8B}} \\
        \hspace{0.5em} GRPO only \scriptsize{(best acc.)}    & 2.370 & \underline{2.053} & \underline{2.773} & 3.657 & 2.713 \\
        \hspace{0.5em} {\mname} \scriptsize{(matched acc.)}  & \textbf{2.658} & \textbf{2.727} & \textbf{4.379} & \underline{3.903} & \textbf{3.417} \\
        \hspace{0.5em} GRPO only \scriptsize{(same step)}    & \underline{2.384} & 2.048 & \underline{2.773} & \textbf{3.911} & \underline{2.779} \\
        \midrule[0.50pt]
        \multicolumn{6}{l}{\textit{R1-Distill-Llama-8B}} \\
        \hspace{0.5em} GRPO only \scriptsize{(best acc.)}    & 1.946 & 1.728 & 2.302 & 3.698 & 2.418 \\
        \hspace{0.5em} {\mname} \scriptsize{(matched acc.)}  & \textbf{2.578} & \textbf{2.984} & \textbf{4.620} & \textbf{4.108} & \textbf{3.572} \\
        \hspace{0.5em} GRPO only \scriptsize{(same step)}    & \underline{2.122} & \underline{1.896} & \underline{2.345} & \underline{3.818} & \underline{2.545} \\
        \bottomrule[1pt]
    \end{tabular}
    \vspace{-5pt}
\end{table}

\begin{table}[t]
    \centering
    \footnotesize
    \caption{\textbf{Reasoning comparison with Claude Opus 4.6.} Accuracy-controlled evaluation of reasoning quality using Claude Opus 4.6 as the judge.}
    \label{table:llm-judge-co4}
    \vspace{-5pt}
    \begin{tabular}{lc*{4}{c}}
        \toprule[1pt]
        \textbf{Model}
            & Relev. $\uparrow$ 
            & Compl. $\uparrow$ 
            & Clarity $\uparrow$ 
            & Fluency $\uparrow$
            & Overall $\uparrow$ \\
        \midrule[0.75pt]
        \multicolumn{6}{l}{\textit{Qwen3-8B}} \\
        \hspace{0.5em} GRPO only \scriptsize{(best acc.)}    & \underline{2.669} & \underline{2.230} & \underline{2.124} & 2.267 & \underline{2.323} \\
        \hspace{0.5em} {\mname} \scriptsize{(matched acc.)}  & \textbf{3.257} & \textbf{3.381} & \textbf{3.581} & \textbf{3.146} & \textbf{3.341} \\
        \hspace{0.5em} GRPO only \scriptsize{(same step)}    & 2.618 & 2.229 & 2.057 & \underline{2.325} & 2.307 \\
        \midrule[0.50pt]
        \multicolumn{6}{l}{\textit{R1-Distill-Llama-8B}} \\
        \hspace{0.5em} GRPO only \scriptsize{(best acc.)}    & 2.300 & 1.870 & 1.813 & 2.277 & 2.065 \\
        \hspace{0.5em} {\mname} \scriptsize{(matched acc.)}  & \textbf{2.905} & \textbf{3.136} & \textbf{2.568} & \textbf{2.983} & \textbf{2.898} \\
        \hspace{0.5em} GRPO only \scriptsize{(same step)}    & \underline{2.416} & \underline{2.016} & \underline{1.907} & \underline{2.388} & \underline{2.182} \\
        \bottomrule[1pt]
    \end{tabular}
    \vspace{-5pt}
\end{table}

\begin{table}[t]
    \centering
    \footnotesize
    \caption{\textbf{Reasoning comparison with Gemini 3 Flash.} Accuracy-controlled evaluation of reasoning quality using Gemini 3 Flash as the judge.}
    \label{table:llm-judge-g3f}
    \vspace{-5pt}
    \begin{tabular}{lc*{4}{c}}
        \toprule[1pt]
        \textbf{Model}
            & Relev. $\uparrow$ 
            & Compl. $\uparrow$ 
            & Clarity $\uparrow$ 
            & Fluency $\uparrow$
            & Overall $\uparrow$ \\
        \midrule[0.75pt]
        \multicolumn{6}{l}{\textit{Qwen3-8B}} \\
        \hspace{0.5em} GRPO only \scriptsize{(best acc.)}    & \underline{2.464} & \underline{2.604} & \underline{2.390} & 3.382 & 2.710 \\
        \hspace{0.5em} {\mname} \scriptsize{(matched acc.)}  & \textbf{2.652} & \textbf{3.610} & \textbf{3.293} & \textbf{4.449} & \textbf{3.501} \\
        \hspace{0.5em} GRPO only \scriptsize{(same step)}    & 2.451 & 2.600 & 2.339 & \underline{3.588} & \underline{2.745} \\
        \midrule[0.50pt]
        \multicolumn{6}{l}{\textit{R1-Distill-Llama-8B}} \\
        \hspace{0.5em} GRPO only \scriptsize{(best acc.)}    & 1.925 & 1.990 & 1.840 & 3.517 & 2.318 \\
        \hspace{0.5em} {\mname} \scriptsize{(matched acc.)}  & \textbf{2.461} & \textbf{3.844} & \textbf{2.970} & \textbf{4.095} & \textbf{3.343} \\
        \hspace{0.5em} GRPO only \scriptsize{(same step)}    & \underline{2.130} & \underline{2.243} & \underline{2.073} & \underline{3.520} & \underline{2.492} \\
        \bottomrule[1pt]
    \end{tabular}
    \vspace{-5pt}
\end{table}

We employ three judge models for LLM-based reasoning evaluation to account for inter-judge variability and individual biases~\citep{kim2026self}.
In addition to reporting average scores across judges, we also provide per-judge results (Tables~\ref{table:llm-judge-4o}--\ref{table:llm-judge-g3f}).
All three judges agree on the overall ranking---{\mname} consistently outperforms both GRPO-only checkpoints---although they differ in calibration and sensitivity across evaluation dimensions.

GPT-4o produces the most moderate scores and shows the strongest clarity signal, with the largest gap between {\mname} and GRPO-only on this dimension.
Fluency scores are relatively compressed and high across all settings, with GPT-4o rarely assigning scores below 2, suggesting some tolerance toward the repetitive patterns often found in degenerate GRPO-only traces.
Qwen3-8B fluency is the one exception where GRPO-only (same step) narrowly exceeds {\mname} (3.911 vs.~3.903), although the difference is negligible.

Claude Opus 4.6 is the strictest judge overall, assigning baseline models scores around 1.8--2.6 across dimensions.
It applies the fluency criterion most aggressively: for Qwen3-8B, 76\% of GRPO-only (best accuracy) traces receive a fluency score of 2 or below, compared to 21\% for {\mname}.
Opus also rarely assigns clarity ratings of 5, indicating a higher bar for fully concrete and logically sound analysis.

Gemini 3 Flash is the most lenient judge, particularly on fluency.
It is also the most bimodal on relevance, with baseline traces scoring very high on checkmate positions but very low on non-mate positions.
This suggests less consistent application of the grounding criterion, with checkmate identification more readily accepted as sufficient justification.
Despite this, Gemini assigns the largest completeness gains to {\mname}.

Taken together, the three judges differ in calibration but agree on the central findings.
Opus and GPT-4o apply the grounding criterion more consistently, whereas Gemini places greater emphasis on structured reasoning and fluency.
Across all judges, {\mname} achieves the highest overall reasoning quality, demonstrating the robustness of structured reasoning with adaptive supervision.

\paragraph{Human agreement with LLM judges.}

To validate our LLM-based reasoning evaluation, we conduct a human study on 50 reasoning traces sampled to span the full range of automated scores.
A human annotator evaluates each trace on the same four dimensions used by the LLM judges using a 1--5 scale.
We then compute the Spearman rank correlation between the human ratings and those assigned by Claude Opus 4.6.
The resulting correlations are 0.89 for relevance, 0.69 for completeness, 0.83 for clarity, and 0.81 for fluency, indicating strong agreement overall.
The relatively lower correlation for completeness likely reflects the greater subjectivity involved in assessing whether a chess analysis is sufficiently thorough.
These results provide additional evidence that LLM-based evaluation is a reliable proxy for human judgments of reasoning quality in our setting.

\section{Additional experimental results}
\label{app:more_eval_results}

\subsection{Comparison with LLM-based process supervision}
\label{app:judge_proc_supervision}

To compare deterministic process supervision with model-based supervision, we train a baseline that uses an LLM judge as the process reward during RL optimization.
Specifically, we use GPT-4o-mini to evaluate each generated reasoning trace according to the same four dimensions: relevance, completeness, clarity, and fluency.
During RL, the model is jointly optimized for prediction accuracy and the four judge-derived process rewards.
We select the checkpoint with the highest held-out prediction accuracy and evaluate reasoning quality using two independent judges, GPT-4o and Gemini 3 Flash.

\begin{table}[t]
    \centering
    \footnotesize
    \caption{\textbf{Comparison with LLM-based process supervision.}
    Reasoning quality of Qwen3-8B trained using GPT-4o-mini as the process reward model compared with {\mname}, evaluated using two independent LLM judges.}
    \label{table:llm-judge-supervision}
    \vspace{-5pt}
    \begin{tabular}{lccccc}
        \toprule[1pt]
        \textbf{Model}
            & Relev. $\uparrow$
            & Compl. $\uparrow$
            & Clarity $\uparrow$
            & Fluency $\uparrow$
            & Overall $\uparrow$ \\
        \midrule[0.75pt]
        \multicolumn{6}{l}{\textit{GPT-4o judge}} \\
        \hspace{0.5em} LLM reward & 2.129 & 1.860 & 2.186 & 2.816 & 2.248 \\
        \hspace{0.5em} {\mname} & \textbf{2.658} & \textbf{2.727} & \textbf{4.379} & \textbf{3.903} & \textbf{3.417} \\
        \midrule[0.50pt]
        \multicolumn{6}{l}{\textit{Gemini 3 Flash judge}} \\
        \hspace{0.5em} LLM reward & 2.461 & 2.453 & 2.104 & 3.579 & 2.649 \\
        \hspace{0.5em} {\mname} & \textbf{2.652} & \textbf{3.610} & \textbf{3.293} & \textbf{4.449} & \textbf{3.501} \\
        \bottomrule[1pt]
    \end{tabular}
    \vspace{-5pt}
\end{table}

Table~\ref{table:llm-judge-supervision} compares LLM-based process supervision with {\mname}.
Across both evaluation judges, {\mname} consistently achieves higher reasoning quality across all dimensions despite relying only on deterministic verifier signals during training.
Qualitatively, we observe that the model trained with LLM supervision often produces informal ``thinking-out-loud'' reasoning traces, suggesting that GPT-4o-mini may reward exploratory or self-questioning writing styles as a proxy for careful reasoning rather than the factual correctness of intermediate claims.
In contrast, {\mname} directly supervises intermediate claims using deterministic verifier signals, encouraging reasoning traces that remain faithful to the underlying state.
LLM-based supervision also incurs approximately 25\% additional wall-clock training time in our implementation due to repeated judge inference during RL optimization.

\subsection{Additional reasoning metrics}
\label{app:more_reason_metrics}

\begin{table}[t]
    \centering
    \footnotesize  %
    \caption{\textbf{Accuracy and reasoning quality comparison.} Additional reasoning quality metrics for Qwen3-8B and DeepSeek-R1-Distill-Llama-8B.}
    \label{table:extra-metrics}
    \vspace{-5pt}
    \begin{tabular}{l*{5}{c}}
        \toprule[1pt]
        \multirow{2.5}{*}{\textbf{Model}} & \multicolumn{4}{c}{\textbf{{Reasoning Quality}}} \\
        \cmidrule(lr){2-5}
         & Capture $\uparrow$ & Check $\uparrow$ & Move Div. $\uparrow$ & Form. Err. $\downarrow$ \\
        \midrule[0.75pt]
        \multicolumn{3}{l}{\textit{Qwen3-8B}} \\
        \hspace{0.3em} SFT only    & \underline{0.399} & 0.651 & \underline{2.193} & \underline{0.003} \\
        \hspace{0.3em} SFT + GRPO  & 0.215 & \underline{0.936} & 1.647 & \underline{0.003} \\
        \hspace{0.3em} {\mname}    & \textbf{0.615} & \textbf{0.988} & \textbf{3.098} & \textbf{0.002} \\
        \midrule[0.50pt]
        \multicolumn{3}{l}{\textit{R1-Distill-Llama-8B}} \\
        \hspace{0.3em} SFT only    & \underline{0.463} & \underline{0.677} & \underline{2.627} & 0.0 \\
        \hspace{0.3em} SFT + GRPO  & 0.120 & 0.441 & 2.388 & 0.0 \\
        \hspace{0.3em} {\mname}    & \textbf{0.523} & \textbf{0.996} & \textbf{4.942} & 0.0 \\
        \bottomrule[1pt]
    \end{tabular}
    \vspace{-5pt}
\end{table}

To provide a more complete picture of reasoning quality, Table~\ref{table:extra-metrics} reports additional metrics beyond those discussed in the main text.
Capture accuracy measures whether the model correctly identifies both the captured piece and its destination square, requiring accurate understanding of the board state and piece interactions.
Check accuracy measures whether the model correctly identifies moves that deliver check, a comparatively simpler task that primarily requires recognizing attacking relationships.
Move diversity (Move Div.) measures the number of distinct candidate moves analyzed, reflecting the breadth of the model's search.
Format error (Form. Err.) reports the fraction of outputs that violate the required reasoning format.

Across both models, outcome-only GRPO improves prediction accuracy but degrades fine-grained reasoning quality, particularly capture accuracy, suggesting weaker grounding in the board state.
In contrast, {\mname} consistently improves all reasoning metrics, achieving higher capture and check accuracy while increasing move diversity, indicating broader and better-grounded analysis.
Check accuracy approaches saturation, and capture accuracy shows substantial gains despite its greater difficulty.
Format errors remain negligible across all methods, with {\mname} yielding a small additional improvement for Qwen3-8B.

\subsection{Analysis of adaptive weighting dynamics}

\begin{figure}[t]
     \centering
     \includegraphics[width=\linewidth]{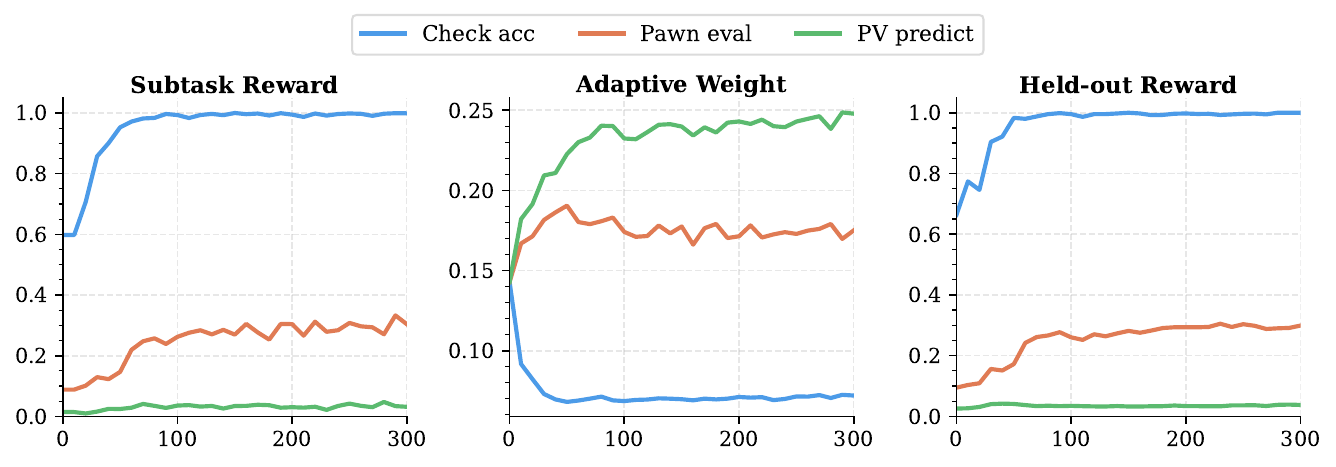}
     \vspace{-15pt}
     \caption{\textbf{Adaptive subtask weighting dynamics.} (\textbf{Left}) Running EMA-based estimates of per-subtask model performance. (\textbf{Center}) Corresponding weights allocated to each subtask that are dynamically adjusted. (\textbf{Right}) Subtask rewards on held-out positions.}
     \label{fig:adapt-demo}
     \vspace{-5pt}
\end{figure}

Figure~\ref{fig:adapt-demo} illustrates how adaptive weighting evolves throughout training as different reasoning subtasks are learned.
Check accuracy, which primarily requires understanding chess rules (i.e., whether a move delivers check), is mastered early: the training reward approaches 1.0 within the first 50 steps, and held-out reward similarly saturates to 1.0.
As a result, the adaptive weight for this subtask drops sharply from the uniform initialization ($1/7 \approx 0.143$) to the minimum floor of 0.05 within the first 20 steps.

Pawn evaluation, which requires estimation of positional advantage, is more challenging but still involves single-position reasoning.
Its training and held-out rewards improve steadily (reaching approximately 0.30 by step 300 on held-out positions), and its weight stabilizes at an intermediate level between the easiest and most difficult subtasks.

PV prediction is the most difficult subtask because it requires predicting a sequence of engine-optimal moves rather than evaluating a single position.
Its training reward remains below 0.05, and held-out performance also improves slowly.
The adaptive mechanism assigns PV prediction the largest weight throughout training (reaching 0.24 by step 300), thereby directing learning toward the most underperforming subtask.

Overall, these dynamics confirm that adaptive weighting behaves as intended: as easier subtasks approach saturation, their weights decrease and optimization progressively shifts toward subtasks with greater remaining headroom.
This adaptive reallocation enables training to focus on the most challenging aspects of reasoning throughout RL.

\subsection{Performance across model scales}

\begin{table}[t]
    \centering
    \footnotesize  %
    \caption{\textbf{Model scale comparison.} Prediction accuracy and reasoning quality metrics
             for Qwen3-8B, Qwen3-4B, and Qwen3-1.7B.}
    \label{table:model-scale}
    \vspace{-5pt}
    \begin{tabular}{l*{7}{c}}
        \toprule[1pt]
        \multirow{2.5}{*}{\textbf{Model}} & \multicolumn{3}{c}{\textbf{{Prediction Accuracy}}} & \multicolumn{3}{c}{\textbf{{Reasoning Quality}}} \\
        \cmidrule(lr){2-4} \cmidrule(lr){5-7}
         & Top-1 Acc. $\uparrow$ & Elo $\uparrow$ & Coverage $\uparrow$ & WR MAE $\downarrow$ & PV Overlap $\uparrow$ & Consist. $\uparrow$ \\
        \midrule[0.75pt]
        \multicolumn{6}{l}{\textit{Qwen3-8B}} \\
        \hspace{0.3em} GRPO only  & 0.507 & 1550 & -- & -- & -- & -- \\
        \hspace{0.3em} SFT only   & 0.212 & 1160 & 0.207 & \underline{0.310} & \underline{0.323} & \underline{0.961} \\
        \hspace{0.3em} SFT + GRPO & \underline{0.534} & \underline{1670} & \underline{0.302} & 0.452 & 0.183 & 0.721 \\
        \hspace{0.3em}  {\mname}  & \textbf{0.536} & \textbf{1706} & \textbf{0.441} & \textbf{0.218} & \textbf{0.376} & \textbf{0.978} \\
        \midrule[0.50pt]
        \multicolumn{6}{l}{\textit{Qwen3-4B}} \\
        \hspace{0.3em} GRPO only  & 0.526 & 1650 & -- & -- & -- & -- \\
        \hspace{0.3em} SFT only   & 0.164 & 1078 & 0.183 & \underline{0.363} & 0.297 & \underline{0.964} \\
        \hspace{0.3em} SFT + GRPO & \textbf{0.546} & \textbf{1663} & \underline{0.410} & 1.0 & \underline{0.321} & 0.0 \\
        \hspace{0.3em} {\mname}   & \underline{0.544} & \underline{1662} & \textbf{0.441} & \textbf{0.238} & \textbf{0.386} & \textbf{0.968} \\
        \midrule[0.50pt]
        \multicolumn{6}{l}{\textit{Qwen3-1.7B}} \\
        \hspace{0.3em} GRPO only  & 0.0 & -- & -- & -- & -- & -- \\
        \hspace{0.3em} SFT only   & 0.128 & 738 & \underline{0.201} & \underline{0.446} & \textbf{0.241} & \underline{0.929} \\
        \hspace{0.3em} SFT + GRPO & \underline{0.514} & \underline{1580} & 0.199 & 0.999 & 0.006 & 0.001 \\
        \hspace{0.3em} {\mname}   & \textbf{0.522} & \textbf{1646} & \textbf{0.410} & \textbf{0.250} & \underline{0.209} & \textbf{0.979} \\
        \bottomrule[1pt]
    \end{tabular}
    \vspace{-5pt}
\end{table}

\begin{table}[t]
    \centering
    \footnotesize  %
    \caption{\textbf{Model scale comparison.} Additional reasoning quality metrics
             for Qwen3-8B, Qwen3-4B, and Qwen3-1.7B.}
    \label{table:model-scale2}
    \vspace{-5pt}
    \begin{tabular}{l*{5}{c}}
        \toprule[1pt]
        \multirow{2.5}{*}{\textbf{Model}} & \multicolumn{4}{c}{\textbf{{Reasoning Quality}}} \\
        \cmidrule(lr){2-5}
         & Capture $\uparrow$ & Check $\uparrow$ & Move Div. $\uparrow$ & Form. Err. $\downarrow$ \\
        \midrule[0.75pt]
        \multicolumn{3}{l}{\textit{Qwen3-8B}} \\
        \hspace{0.3em} SFT only    & \underline{0.399} & 0.651 & \underline{2.193} & \underline{0.003} \\
        \hspace{0.3em} SFT + GRPO  & 0.215 & \underline{0.936} & 1.647 & \underline{0.003} \\
        \hspace{0.3em} {\mname}    & \textbf{0.615} & \textbf{0.988} & \textbf{3.098} & \textbf{0.002} \\
        \midrule[0.50pt]
        \multicolumn{3}{l}{\textit{Qwen3-4B}} \\
        \hspace{0.3em} SFT only   & \underline{0.369} & \underline{0.438} & 2.067 & \underline{0.004} \\
        \hspace{0.3em} SFT + GRPO & 0.072 & 0.360 & \underline{2.751} & 0.005 \\
        \hspace{0.3em} {\mname}   & \textbf{0.614} & \textbf{0.996} & \textbf{3.243} & \textbf{0.0} \\
        \midrule[0.50pt]
        \multicolumn{3}{l}{\textit{Qwen3-1.7B}} \\
        \hspace{0.3em} SFT only   & 0.030 & 0.024 & \underline{2.661} & 0.035 \\
        \hspace{0.3em} SFT + GRPO & \underline{0.067} & \underline{0.045} & 1.067 & \underline{0.009} \\
        \hspace{0.3em} {\mname}   & \textbf{0.255} & \textbf{0.941} & \textbf{2.885} & \textbf{0.002} \\
        \bottomrule[1pt]
    \end{tabular}
    \vspace{-5pt}
\end{table}

We next evaluate whether the main findings hold consistently across model scales.
Tables~\ref{table:model-scale} and~\ref{table:model-scale2} report results for Qwen3-8B, 4B, and 1.7B models.
The core findings at the 8B scale replicate consistently across smaller models.
At 4B and 1.7B, SFT + GRPO drives win-rate MAE close to its maximum while reducing consistency to nearly zero, confirming that outcome-only RL systematically erodes reasoning quality.
Moreover, this degradation emerges earlier at smaller scales: win-rate MAE exceeds 0.90 by step 200 for 8B, step 100 for 4B, and step 80 for 1.7B, suggesting that smaller models are more susceptible to reasoning collapse under outcome-only supervision.
In contrast, {\mname} maintains near-perfect consistency (0.97--0.98) while reducing win-rate MAE well below the SFT baseline (0.24 vs.~0.36 at 4B; 0.25 vs.~0.45 at 1.7B).
Check accuracy reaches 0.94--0.99 under {\mname}, compared to 0.05--0.36 under SFT + GRPO, with similar gains in capture accuracy (0.26--0.61 vs.~0.06--0.07).

Across all model scales, {\mname} matches or exceeds SFT + GRPO in both top-1 accuracy and Elo rating.
At 8B and 1.7B, it achieves higher Elo ratings (1706 vs.~1670 for 8B; 1646 vs.~1580 for 1.7B), indicating that improved reasoning translates to stronger overall performance.
Interestingly, Qwen3-4B achieves slightly higher top-1 accuracy than Qwen3-8B despite starting from a weaker SFT baseline (0.16 vs.~0.21).
However, the 8B model attains higher Elo and stronger reasoning metrics, suggesting that its additional capacity is devoted to improving overall move quality and reasoning rather than top-1 accuracy within the training horizon considered.
Outcome-only GRPO trained from scratch fails to converge at the 1.7B scale (0.0 accuracy and 1.0 format error throughout training), whereas SFT initialization enables stable learning.
This further highlights the importance of a structured reasoning prior, particularly for smaller models.

\subsection{Effect of reasoning budget on learned patterns}
\label{app:reason_budget}

\begin{figure}[t]
     \centering
     \begin{subfigure}[b]{0.48\textwidth}
         \centering
         \includegraphics[width=\textwidth]{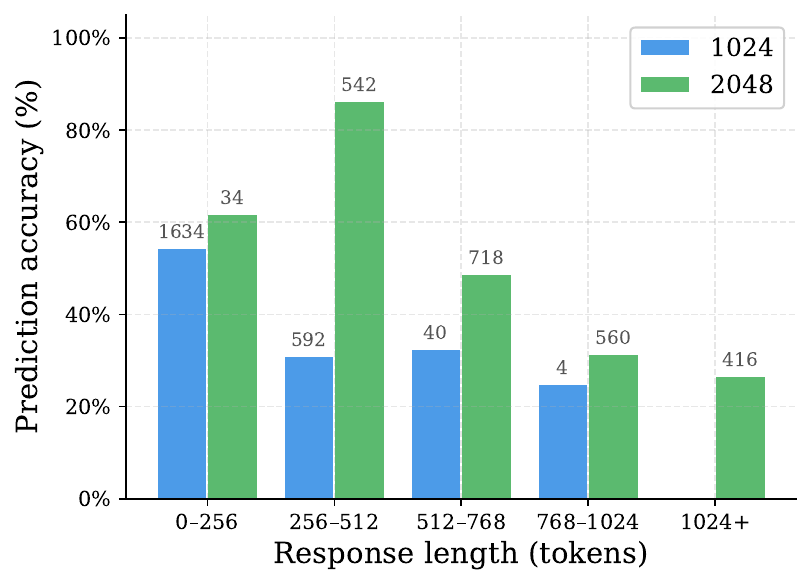}
     \end{subfigure}
     \hfill
     \begin{subfigure}[b]{0.48\textwidth}
         \centering
         \includegraphics[width=\textwidth]{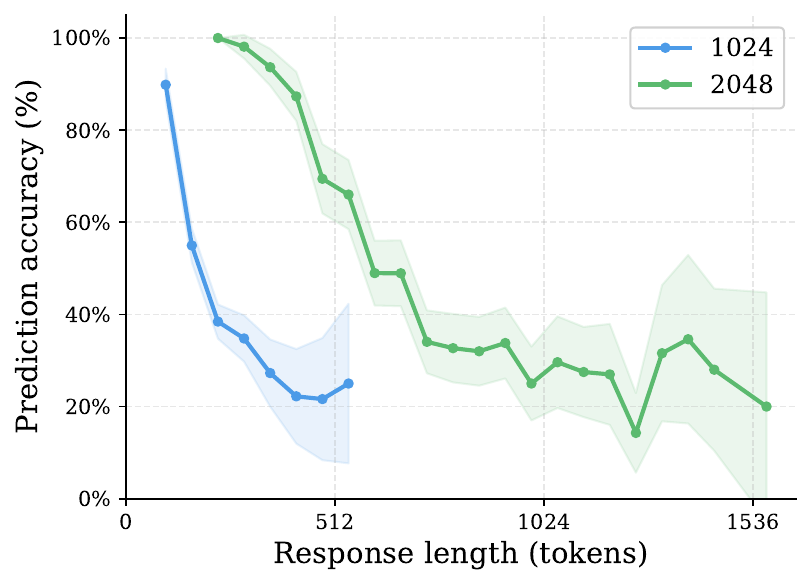}
     \end{subfigure}
     \vspace{-5pt}
     \caption{\textbf{Accuracy vs.\ reasoning length under different budgets.} Short traces achieve high accuracy, while longer traces correlate with substantially lower accuracy. Distributions show that larger budgets induce longer, more dispersed reasoning traces.}
     \label{fig:len_acc_comp}
     \vspace{-5pt}
\end{figure}

Figure~\ref{fig:len_acc_comp} illustrates how prediction accuracy varies with response length (measured in tokens) for models trained with different reasoning budgets (maximum lengths of 1024 and 2048).
The left plot groups traces into five coarse bins and reports the mean accuracy within each.
Sample counts above each bar show that the 1024-token configuration concentrates most traces in the shortest bin (0--256 tokens), whereas the 2048-token configuration distributes traces more evenly across the full range.
This difference indicates that models converge to distinct reasoning patterns depending on the available reasoning budget.

The right plot provides a finer-grained view using 64-token bins with 95\% confidence intervals.
Both plots reveal a consistent negative relationship between trace length and prediction accuracy: traces shorter than 256 tokens achieve over 80\% accuracy, whereas traces longer than 1024 tokens achieve below 30\%, with Spearman correlations of $\rho=-0.36$ (1024) and $\rho=-0.44$ (2048).
Together, these results suggest that, in the absence of structured supervision, longer reasoning traces often reflect confusion than productive deliberation.

\subsection{Faithfulness and robustness of learned reasoning}

\begin{table}[t]
    \centering
    \footnotesize
    \caption{\textbf{Faithfulness and robustness of learned reasoning.}
    (\textbf{Left}) Sensitivity of model predictions to perturbations of intermediate reasoning.
    (\textbf{Right}) Prediction accuracy under alternative prompting schemes.
    Results are averaged over three random seeds.}
    \label{table:faithfulness}
    \vspace{-5pt}
    \begin{subtable}[t]{0.42\linewidth}
        \centering
        \caption{Reasoning intervention}
        \label{table:reasoning_intervention}
        \begin{tabular}{lcc}
            \toprule[1pt]
            \textbf{Model}
                & Conclusion
                & Win rate \\
            \midrule[0.75pt]
            SFT + GRPO
                & $0.578$
                & $0.415$ \\
            {\mname}
                & $\mathbf{0.836}$
                & $\mathbf{0.936}$ \\
            \bottomrule[1pt]
        \end{tabular}
    \end{subtable}
    \hfill
    \begin{subtable}[t]{0.55\linewidth}
        \centering
        \caption{Prompt robustness}
        \label{table:prompt_robustness}
        \begin{tabular}{lccc}
            \toprule[1pt]
            \textbf{Model}
                & Original
                & No Struct.
                & Free Form \\
            \midrule[0.75pt]
            SFT + GRPO
                & $0.518$
                & $0.504$
                & $0.477$ \\
            {\mname}
                & $\mathbf{0.542}$
                & $\mathbf{0.522}$
                & $\mathbf{0.494}$ \\
            \bottomrule[1pt]
        \end{tabular}
    \end{subtable}
    \vspace{-5pt}
\end{table}

To evaluate the robustness of the learned reasoning, we perform two additional analyses on held-out positions.
First, we evaluate the sensitivity of model predictions to perturbations of intermediate reasoning.
Specifically, we consider two interventions: (1) replacing the move stated in the conclusion with another candidate move, and (2) perturbing the predicted win-rate ordering among candidate moves.
For each intervention, we measure the fraction of positions for which the final prediction changes, averaged over three seeds.
Table~\ref{table:reasoning_intervention} shows that perturbing intermediate reasoning produces substantially larger changes in the predictions of {\mname} than of SFT + GRPO.
In particular, corrupting predicted win-rate estimates changes the final prediction for over 93\% of positions for {\mname}, compared to only 41.5\% for SFT + GRPO.
Qualitatively, we observe that SFT + GRPO frequently predicts nearly identical win-rates across moves or omits them entirely, indicating that these quantities are only weakly coupled to the final prediction.
These results suggest that predictions produced by {\mname} depend more strongly on the generated intermediate reasoning.

We also evaluate whether the learned reasoning remains robust under alternative prompting schemes.
In addition to the original prompt, we considered two variations: (a) removing reminders about basic chess structure (``No Structure''), and (b) allowing completely free-form reasoning (``Free Form'').
As shown in Table~\ref{table:prompt_robustness}, both models experience reduced prediction accuracy under these prompt variations. 
However, the degradation is consistently smaller for {\mname}, suggesting that the structured reasoning learned through process supervision generalizes beyond the specific prompting format used during training.

\subsection{Qualitative analysis}

\begin{figure}[t]
     \centering
     \begin{subfigure}[b]{0.32\textwidth}
         \centering
         \includegraphics[width=\textwidth]{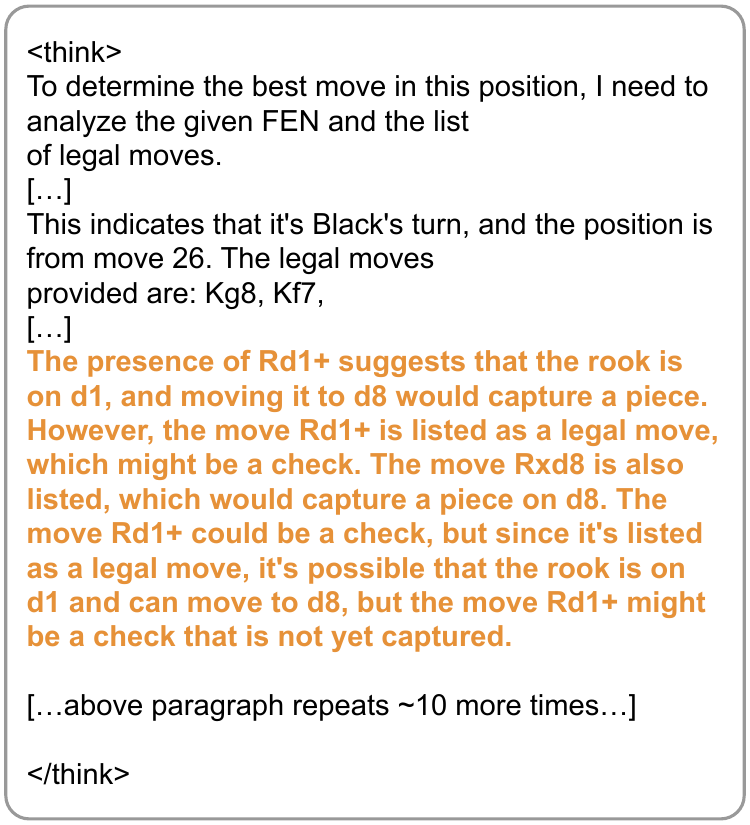}
         \caption{GRPO only}
     \end{subfigure}
     \hfill
     \begin{subfigure}[b]{0.32\textwidth}
         \centering
         \includegraphics[width=\textwidth]{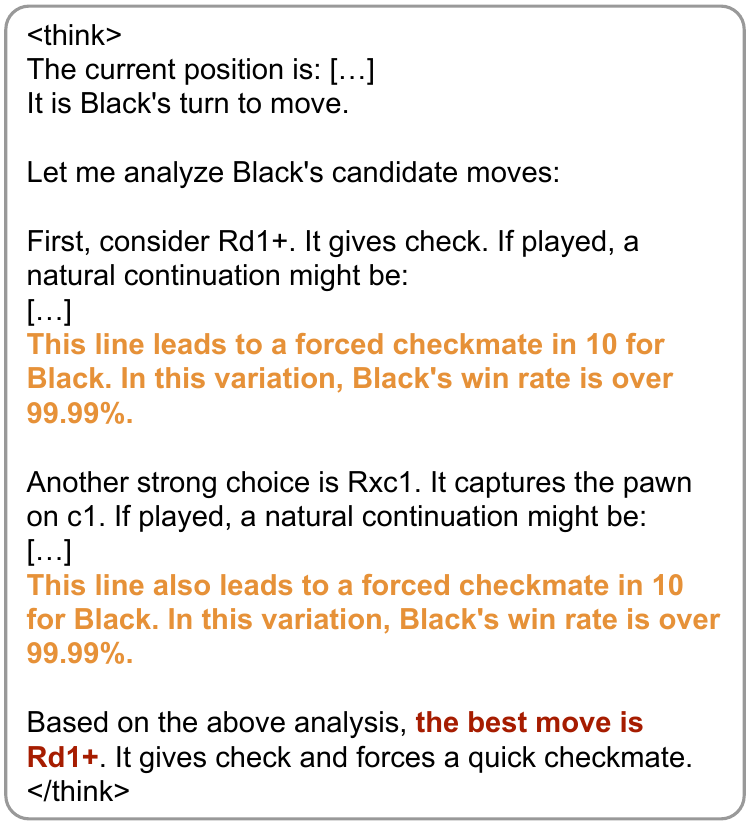}
         \caption{SFT + GRPO}
     \end{subfigure}
     \hfill
     \begin{subfigure}[b]{0.32\textwidth}
         \centering
         \includegraphics[width=\textwidth]{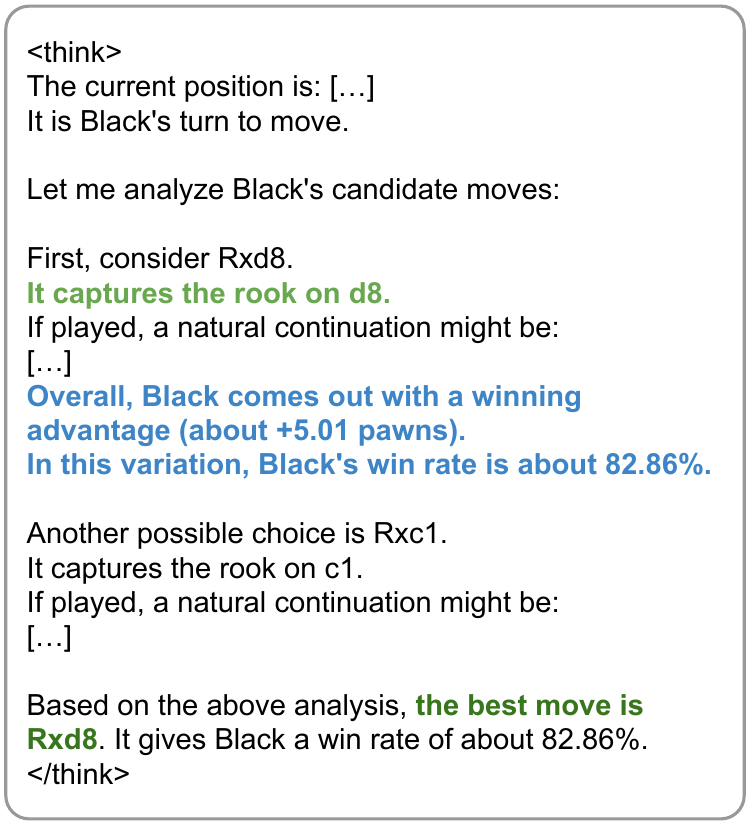}
         \caption{{\mname}}
     \end{subfigure}
     \vspace{-5pt}
     \caption{\textbf{Qualitative reasoning comparison.} GRPO exhibits repetition and non-termination, while SFT + GRPO produces superficial, overconfident reasoning. {\mname} yields grounded, calibrated analysis with correct move selection.}
     \label{fig:trace_comp_app1}
     \vspace{-5pt}
\end{figure}

We provide additional qualitative examples illustrating the reasoning patterns learned under different training settings.
Figure~\ref{fig:trace_comp_app1} contrasts baseline models with {\mname}, highlighting common failure modes such as repetition, looping, and superficial heuristic reasoning.
Under GRPO alone, the model identifies two plausible candidates (\texttt{Rxd8} and \texttt{Rd1+}) but fails to resolve between them, entering a verbatim loop that repeats the same paragraph multiple times before terminating without producing an answer.
With SFT + GRPO, the model produces more structured traces but remains largely superficial: it selects the checking move \texttt{Rd1+} and immediately claims a forced mate in 10 with ``over 99.99\%'' win rate, despite no such continuation existing.
Alternative candidates receive nearly identical boilerplate analysis, differing only in the first move, indicating little position-specific evaluation.
In contrast, {\mname} produces more coherent reasoning by evaluating multiple distinct candidates (\texttt{Rxd8}, \texttt{Rxc1}, and \texttt{Rd1+}), assigning plausible win-rate estimates ($\sim$82\%), and correctly identifying \texttt{Rxd8} as the strongest move.
The reasoning is non-repetitive and grounded in the position, with only minor inaccuracies in deeper continuations.

\begin{figure}[t]
     \centering
     \begin{subfigure}[b]{0.32\textwidth}
         \centering
         \includegraphics[width=\textwidth]{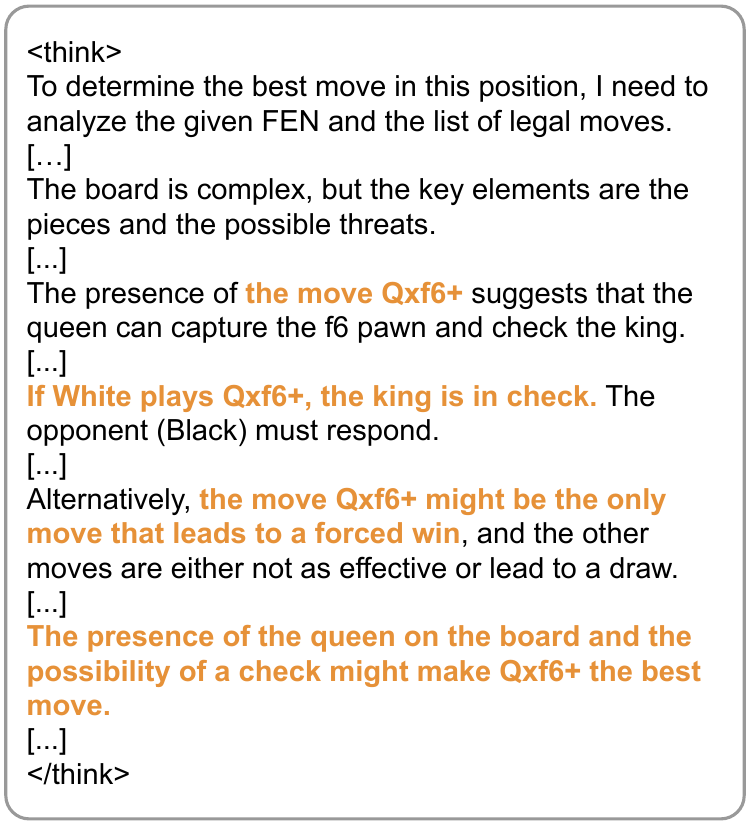}
         \caption{GRPO only}
     \end{subfigure}
     \hfill
     \begin{subfigure}[b]{0.32\textwidth}
         \centering
         \includegraphics[width=\textwidth]{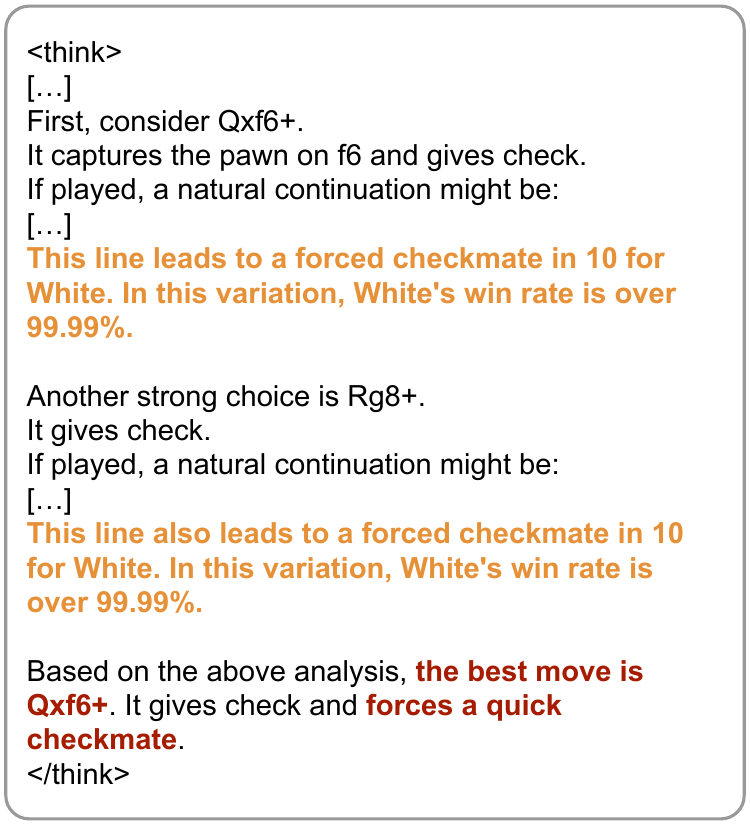}
         \caption{SFT + GRPO}
     \end{subfigure}
     \hfill
     \begin{subfigure}[b]{0.32\textwidth}
         \centering
         \includegraphics[width=\textwidth]{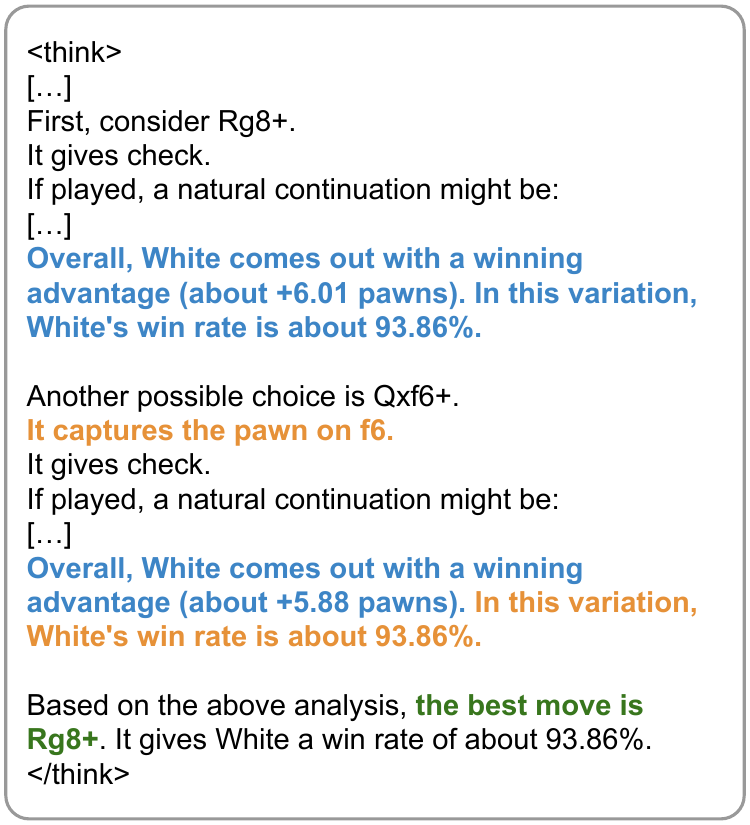}
         \caption{{\mname}}
     \end{subfigure}
     \vspace{-5pt}
     \caption{\textbf{Qualitative reasoning comparison.} Baselines fixate on \texttt{Qxf6+} and produce superficial or templated analysis, while {\mname} correctly prioritizes \texttt{Rg8+} based on its forcing continuation, yielding more grounded reasoning.}
     \label{fig:trace_comp_app2}
     \vspace{-5pt}
\end{figure}

Figure~\ref{fig:trace_comp_app2} illustrates similar differences in reasoning quality.
Under GRPO, the model fixates on \texttt{Qxf6+} and discusses the move in a circuitous manner without meaningful analysis.
With SFT + GRPO, the model again fixates on \texttt{Qxf6+} because it gives check, fabricating a ``forced mate in 10'' for both \texttt{Qxf6+} and \texttt{Rg8+}, with nearly identical analysis.
In contrast, {\mname} correctly identifies \texttt{Rg8+} as the stronger initial move, recognizing that it forces a recapture by the king before enabling \texttt{Qxf6+} under more favorable conditions.
The resulting reasoning is more structured and grounded in the position, although the model still makes a minor error by incorrectly identifying a capture in one continuation.

\begin{figure}[t]
     \centering
     \begin{subfigure}[b]{0.32\textwidth}
         \centering
         \includegraphics[width=\textwidth]{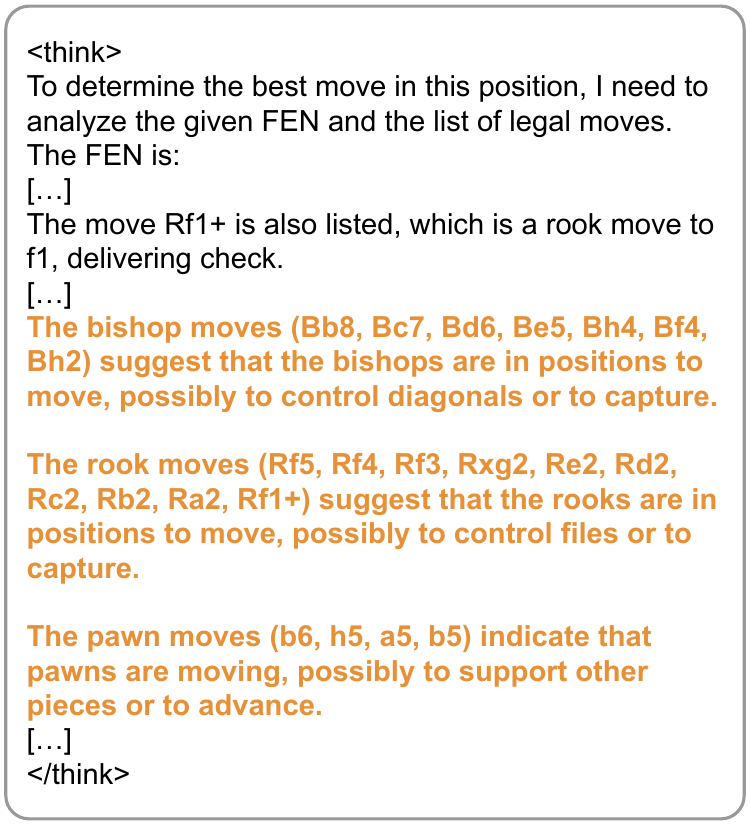}
         \caption{GRPO only}
     \end{subfigure}
     \hfill
     \begin{subfigure}[b]{0.32\textwidth}
         \centering
         \includegraphics[width=\textwidth]{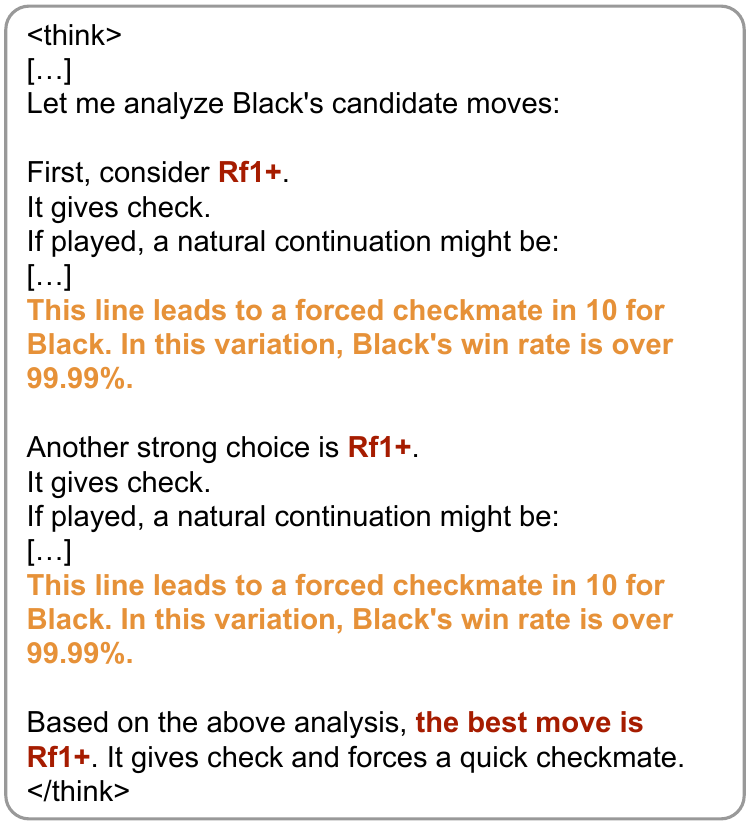}
         \caption{SFT + GRPO}
     \end{subfigure}
     \hfill
     \begin{subfigure}[b]{0.32\textwidth}
         \centering
         \includegraphics[width=\textwidth]{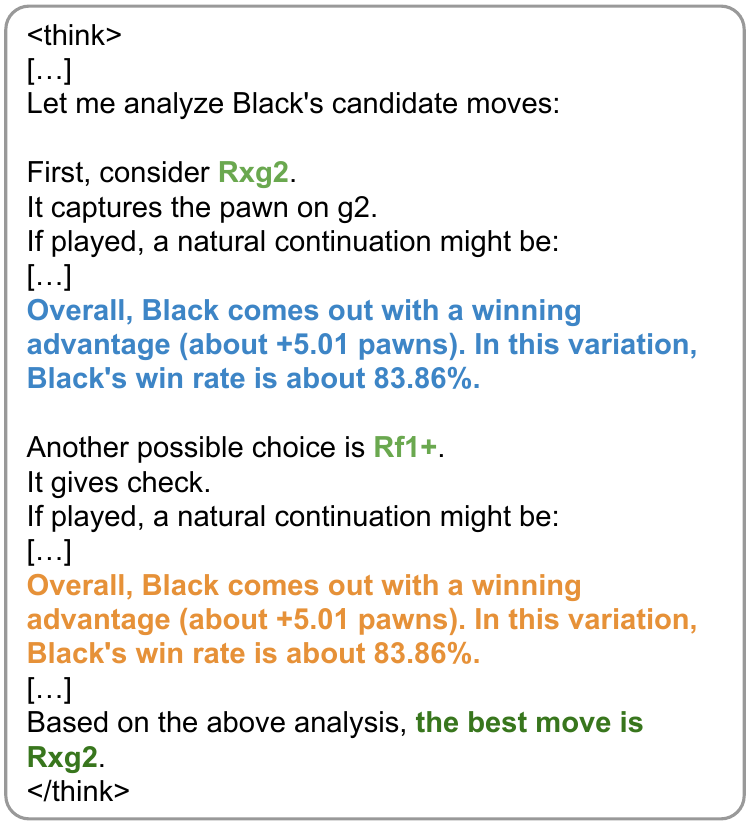}
         \caption{{\mname}}
     \end{subfigure}
     \vspace{-5pt}
     \caption{\textbf{Qualitative reasoning comparison.} Baseline models perform minimal analysis or rely on shortcut heuristics, repeating checking moves without position-specific evaluation.}
     \label{fig:trace_comp_app3}
     \vspace{-5pt}
\end{figure}

Figure~\ref{fig:trace_comp_app3} shows examples in which the baseline models either perform minimal analysis or repeatedly prioritize checking moves.
Under GRPO, the model provides only superficial analysis, grouping moves by piece type and reciting generic chess principles without engaging the specific position.
SFT + GRPO similarly collapses to the heuristic of ``give check $\rightarrow$ claim forced mate,'' listing \texttt{Rf1+} twice as its top candidate with identical continuation lines.
In contrast, {\mname} produces more grounded evaluations, correctly prioritizing \texttt{Rxg2} for its immediate material gain while treating \texttt{Rf1+} as a secondary candidate.
Although the reasoning is substantially more calibrated and position-aware, the model still exhibits minor flaws, such as assigning similar evaluations to distinct moves, likely reflecting the difficulty of learning fine-grained chess reasoning without extensive domain-specific training.

\subsection{Additional GSM8K results}

\paragraph{Contribution of individual verifier signals.}

\begin{table}[t]
    \centering
    \footnotesize
    \caption{\textbf{Contribution of individual verifier signals on GSM8K.}
    Each variant is trained with a single verifier signal or the full objective. Results are averaged over three seeds.}
    \label{table:gsm8k_reward_ablation}
    \vspace{-5pt}
    \begin{tabular}{lccc}
        \toprule[1pt]
        \textbf{Method}
            & Accuracy $\uparrow$
            & Step Arith. $\uparrow$
            & Ans. Cons. $\uparrow$ \\
        \midrule[0.75pt]
        SFT baseline
            & $0.446$
            & $0.638$
            & $0.821$ \\
        Step arith. only
            & \underline{0.753}
            & \textbf{0.906}
            & $0.817$ \\
        Ans. cons. only
            & $0.752$
            & $0.863$
            & \underline{0.865} \\
        {\mname}
            & \textbf{0.763}
            & \underline{0.905}
            & \textbf{0.889} \\
        \bottomrule[1pt]
    \end{tabular}
    \vspace{-5pt}
\end{table}

To better understand the contribution of individual verifier signals, we perform an ablation on GSM8K by training models with (1) step-arithmetic supervision only, (2) answer-consistency supervision only, and (3) the full {\mname} objective combining both rewards.
Table~\ref{table:gsm8k_reward_ablation} shows that both individual rewards substantially improve prediction accuracy over the SFT baseline.
As expected, each reward primarily improves the aspect of reasoning that it directly supervises.
Step-arithmetic supervision achieves the highest arithmetic correctness among the single-reward variants (0.906), whereas answer-consistency supervision produces the highest answer consistency (0.865).
However, neither reward alone performs strongly across all evaluation metrics.
In contrast, jointly optimizing both verifier signals provides the strongest overall trade-off, attaining the highest prediction accuracy (0.763), matching the best step-arithmetic correctness (0.905), and producing the highest answer consistency (0.889).
These results suggest that the verifier signals capture complementary aspects of reasoning quality and that jointly optimizing them yields more balanced and reliable reasoning behavior than optimizing either process reward in isolation.

\paragraph{Generalization to non-reasoning models.}

\begin{table}[t]
    \centering
    \footnotesize
    \caption{\textbf{Generalization to a non-reasoning language model.}
    Results on GSM8K using Qwen3-0.6B-Base, the non-reasoning counterpart of Qwen3-0.6B. The qualitative findings mirror those observed for reasoning models.}
    \label{table:gsm8k_base}
    \vspace{-5pt}
    \begin{tabular}{lccc}
        \toprule[1pt]
        \textbf{Method}
            & Accuracy $\uparrow$
            & Step Arith. $\uparrow$
            & Ans. Cons. $\uparrow$ \\
        \midrule[0.75pt]
        SFT baseline
            & 0.528
            & \underline{0.669}
            & \underline{0.851} \\
        SFT + GRPO
            & \underline{0.721}
            & 0.618
            & 0.731 \\
        {\mname}
            & \textbf{0.722}
            & \textbf{0.974}
            & \textbf{0.958} \\
        \bottomrule[1pt]
    \end{tabular}
    \vspace{-5pt}
\end{table}

Although our main experiments focus on reasoning models, we also evaluate whether {\mname} extends to non-reasoning language models.
We perform this study on GSM8K using Qwen3-0.6B-Base, the non-reasoning counterpart of Qwen3-0.6B.
Table~\ref{table:gsm8k_base} shows that the same qualitative trends extend to the non-reasoning setting.
Outcome-only RL substantially improves prediction accuracy over the SFT baseline (0.721 vs.~0.528), but degrades both step arithmetic (0.618 vs.~0.669) and answer consistency (0.731 vs.~0.851).
In contrast, {\mname} achieves comparable prediction accuracy (0.722) while dramatically improving reasoning quality, reaching 0.974 step arithmetic and 0.958 answer consistency.
Although the non-reasoning model initially produces less consistent reasoning formats than its reasoning counterpart, reliable extraction and verification remain feasible with slightly more permissive parsing rules.
Overall, these results suggest that deterministic process supervision generalizes beyond reasoning language models and can likewise improve the reasoning quality of pretrained models.

\section{Prompt details}
\label{app:prompts}

\subsection{Chess reasoning}
\label{app:chess-prompts}

For the main chess reasoning task, we adopt the prompt from Chess-R1~\citep{hwang2025can}, which instructs the model to analyze key candidate moves in its reasoning and conclude with a predicted optimal move.
We also include a brief summary of the chess rules.

\begin{tcolorbox}[title=Chess reasoning instructions, colback=gray!5, colframe=black!50, width=\textwidth, breakable]
\scriptsize  %
\begin{lstlisting}[breaklines, breakindent=0pt]
You are a helpful assistant who plays chess professionally. First, think through the reasoning process internally and then provide the user with the best move. The reasoning process and the answer must be enclosed within <think> </think> and <answer> </answer> tags, respectively.

The reasoning process should describe how you analyze the position and decide on the best move, including:
- A strategic evaluation of the position.
- A comparison of key candidate moves.
- For each candidate, consider the opponent's likely response and outcome.
- Conclude with a clear justification for the final choice.

The answer must be in SAN notation, restricted to the moving piece and destination square (e.g., Nf3, Rxf2, c5). Now, the user provides the board in FEN format, a list of legal moves for the given board.
After analyzing the position, clearly state the best move in SAN notation within <answer> </answer> tags. i.e., <answer> Nf3 </answer>.

Reminder of chess rules:
- Bishops move diagonally.
- Rooks move horizontally or vertically.
- Knights jump in an L-shape.
- Queens combine rook and bishop movements.
- Kings move one square in any direction.
- Pawns move forward, capture diagonally, and can promote.

Current board in FEN: <board>
Legal moves: <legal moves>
\end{lstlisting}
\end{tcolorbox}

\subsection{Relevance evaluation}

\begin{tcolorbox}[title=Relevance evaluation instructions, colback=gray!5, colframe=black!50, width=\textwidth, breakable]
\scriptsize  %
\begin{lstlisting}[breaklines, breakindent=0pt]
You are an expert chess analyst evaluating the reasoning trace of an AI chess assistant that was asked to find the best move in a given position.

Your task is to rate the reasoning on **Relevance** (1-5).

### Definition
Relevance measures two things jointly:
1. **Candidate selection**: whether the moves analyzed are reasonable to consider, using the engine summary as a reference for which candidates are meaningful.
2. **Analytical grounding**: whether the trace provides position-specific justification for its candidates -- a concrete continuation, a tactical observation, or a move-specific evaluation. Generic statements that could apply to any position ("this is a check", "this ends the game") without position-specific follow-through do not count.

Do not assess the numerical accuracy of evaluations or win rates, or the optimality of continuations. In positions where the engine's top move is a checkmate, analyzing additional checks or captures as candidates is not penalized.

### Scoring Guide
5 - Top move present with position-specific justification; all other candidates reasonable and grounded.
4 - Top move present with some position-specific justification; minor gaps in grounding or at most one unreasonable candidate.
3 - Top move present but justification is largely generic, or top move absent but all candidates are grounded.
2 - Top move absent or present only as a label with no position-specific justification; most analysis generic.
1 - Almost entirely symbol recognition, generic filler, or hopeless moves.

### Instructions
1. Use the engine summary to identify the top candidates.
2. Identify which moves the trace analyzes and whether the top move is present.
3. For each candidate, assess whether the justification is position-specific or merely generic.
4. Assess whether additional candidates are reasonable, regardless of order.
5. Respond: {"score": <int 1-5>, "justification": "<1-2 sentence explanation>"}
\end{lstlisting}
\end{tcolorbox}

\subsection{Completeness evaluation}

\begin{tcolorbox}[title=Completeness evaluation instructions, colback=gray!5, colframe=black!50, width=\textwidth, breakable]
\scriptsize  %
\begin{lstlisting}[breaklines, breakindent=0pt]
You are an expert chess analyst evaluating the reasoning trace of an AI chess assistant that was asked to find the best move in a given position.

Your task is to rate the reasoning on **Completeness** (1-5).

### Definition
Completeness measures whether the reasoning follows through on the analysis it begins. A complete trace introduces multiple candidate moves, provides evaluations, continuations, and/or other analyses for each, and arrives at a supported conclusion. A trace that introduces a candidate and then abandons it without analysis is incomplete.

Numerical accuracy of evaluations is not assessed -- only whether each introduced candidate receives some logical analysis.

### Scoring Guide
5 - Multiple candidates introduced and evaluated with logical justifications; clear supported conclusion.
4 - Nearly all candidates addressed; one may be underdeveloped.
3 - Some candidates addressed but notable gaps remain.
2 - Only one candidate discussed with limited analysis; conclusion unsupported.
1 - Superficial or nearly empty; no real analysis completed.

### Instructions
1. Use the engine summary to orient yourself on the position.
2. Identify all candidate moves introduced in the trace.
3. Assess whether each is followed through with some analysis.
4. Assess whether the conclusion is supported.
5. Respond: {"score": <int 1-5>, "justification": "<1-2 sentence explanation>"}
\end{lstlisting}
\end{tcolorbox}

\subsection{Clarity evaluation}

\begin{tcolorbox}[title=Clarity evaluation instructions, colback=gray!5, colframe=black!50, width=\textwidth, breakable]
\scriptsize  %
\begin{lstlisting}[breaklines, breakindent=0pt]
You are an expert chess analyst evaluating the reasoning trace of an AI chess assistant that was asked to find the best move in a given position.

Your task is to rate the reasoning on **Clarity** (1-5).

### Definition
Clarity measures whether the reasoning is specific, precise, and unambiguous. Concrete evidence -- named moves, specific continuations, centipawn evaluations, and win rates tied to specific lines -- scores higher than vague or hand-wavy assertions such as "this improves the position" without explaining how.

### Scoring Guide
5 - Every claim is specific and concrete, with precise references to moves, lines, and/or evaluations.
4 - Mostly specific; occasional minor vagueness.
3 - A mix of concrete and vague statements.
2 - Predominantly vague or hand-wavy; few concrete references.
1 - Entirely vague or incoherent.

### Instructions
1. Read the reasoning trace.
2. Assess how specific and precise each analytical claim is.
3. Respond with a JSON object: {"score": <int 1-5>, "justification": "<1-2 sentence explanation>"}
\end{lstlisting}
\end{tcolorbox}

\subsection{Fluency evaluation}

\begin{tcolorbox}[title=Fluency evaluation instructions, colback=gray!5, colframe=black!50, width=\textwidth, breakable]
\scriptsize  %
\begin{lstlisting}[breaklines, breakindent=0pt]
You are a language-quality evaluator assessing the reasoning trace of an AI chess assistant.

Your task is to rate the reasoning on **Fluency** (1-5).

### Definition
Fluency measures how well the text is written: grammatical correctness, sentence structure, logical flow, and coherent transitions between ideas. This criterion is about language quality, not chess correctness.

Repetitive or circular reasoning that revisits the same conclusion without new analysis should lower the score.
Consistent structural formatting across multiple candidates is not penalized; only repetition that fails to add new analytical content should lower the score.

### Scoring Guide
5 - Perfectly fluent, well-organized, and reads naturally.
4 - Minor imperfections that do not hinder understanding.
3 - Noticeable grammatical or organizational issues but still understandable.
2 - Frequent errors or circular reasoning that impede comprehension.
1 - Largely unreadable or incoherent.

### Instructions
1. Read the reasoning trace.
2. Evaluate grammar, sentence structure, and organization.
3. Respond with a JSON object: {"score": <int 1-5>, "justification": "<1-2 sentence explanation>"}
\end{lstlisting}
\end{tcolorbox}

\end{document}